\documentclass[final,authoryear,5p,times]{elsarticle}

\usepackage{lineno}

\usepackage{graphicx}
\usepackage{amssymb}
\usepackage{amsmath}
\usepackage{url}
\usepackage{color}
\usepackage{bm}
\usepackage{subfigure}
\usepackage{algorithmic}
\usepackage{algorithm}
\usepackage{siunitx}

\newcommand{\argmax}{\operatornamewithlimits{argmax}}
\newcommand{\argmin}{\operatornamewithlimits{argmin}}

\renewcommand{\algorithmicrequire}{\textbf{Input:}}
\renewcommand{\algorithmicensure}{\textbf{Output:}}

\journal{Preprint}

\begin{document}

\begin{frontmatter}

\title{Joint registration and synthesis using a probabilistic model for alignment of MRI and histological sections}

\author[tig]{Juan Eugenio Iglesias}
\author[tig]{Marc Modat}
\author[weiss]{Lo\"{i}c Peter}
\author[martinos]{Allison Stevens}
\author[tig]{Roberto Annunziata}
\author[weiss]{Tom Vercauteren}
\author[allen]{Ed Lein}
\author[martinos,mit]{Bruce Fischl}
\author[weiss]{Sebastien Ourselin}
\author{for the Alzheimer's Disease Neuroimaging Initiative\fnref{adniBlob}}
\fntext[adniBlob]{Data used in preparation of this article were obtained from the Alzheimer's Disease Neuroimaging Initiative (ADNI) database (\url{http://adni.loni.usc.edu}). As such, the investigators within the ADNI contributed to the design and implementation of ADNI and/or provided data but did not participate in analysis or writing of this report. A complete listing of ADNI investigators can be found at: \url{adni.loni.usc.edu/wp-
content/uploads/how_to_apply/ADNI_Acknowledgement_List.pdf}.}

\address[tig]{Translational Imaging Group, University College London, UK}
\address[weiss]{Wellcome EPSRC Centre for Interventional and Surgical Sciences (WEISS), University College London, UK}
\address[martinos]{Martinos Center for Biomedical Imaging, Harvard Medical School and Massachusetts General Hospital, USA}
\address[allen]{Allen Institute for Brain Science, USA}
\address[mit]{Computer Science and AI lab, Massachusetts Institute of Technology, USA}

\begin{abstract}

Nonlinear registration of 2D  histological sections with corresponding  slices of MRI data is a  critical step of  3D histology reconstruction algorithms. This registration is difficult due to the large differences in image contrast and resolution, as well as the complex nonrigid deformations and artefacts produced when sectioning the sample and mounting it on the glass slide. It has been shown in brain MRI registration that better spatial alignment across modalities can be obtained by synthesizing one modality from the other and then using intra-modality registration metrics, rather than by using information theory based metrics to solve the problem directly. However, such an approach typically requires a database of aligned images from the two modalities, which is very difficult to obtain for histology and MRI. 

Here, we overcome this limitation with a probabilistic method that simultaneously solves for deformable registration and synthesis directly on the target images, without requiring any training data. The method is based on a  probabilistic model  in which the MRI slice is assumed to be a contrast-warped, spatially deformed version of the histological section. We use approximate Bayesian inference to iteratively refine the probabilistic estimate of the synthesis  and the registration,  while accounting for each other's uncertainty. Moreover, manually placed landmarks can be seamlessly integrated in the framework for increased performance and robustness. 

Experiments on a synthetic dataset of MRI slices show that, compared with mutual information based registration, the proposed method makes it possible to use a much more flexible deformation model in the registration  to improve its accuracy, without compromising robustness. Moreover, our framework also exploits information in manually placed landmarks more efficiently than MI, since landmarks inform both synthesis and registration -- as opposed to registration alone. Finally, we show qualitative results on the publicly available Allen atlas, in which the proposed method provides a clear improvement over mutual information based registration.

\end{abstract}


\end{frontmatter}



\section{Introduction}

\subsection{Motivation: human brain atlases}

Histology is the study of tissue microanatomy. Histological analysis involves cutting a wax-embedded or frozen block of tissue into very thin sections (in the order of 10 microns), which are subsequently stained, mounted on glass slides, and examined under the microscope. Using different types of stains, different  microscopic structures can be enhanced and studied. Moreover, mounted sections can be digitised at high resolution -- in the order of a micron. Digital histological sections not only enable digital pathology in a clinical setting, but also open the door to an array of image analysis applications.

A promising application of digital histology is the construction of high resolution computational atlases of the human brain. Such atlases have traditionally been built using MRI scans and/or associated manual segmentations, depending on whether they describe image intensities, neuroanatomical label probabilities, or both. Examples include: the MNI atlas \citep{evans19933d,collins1994automatic}, the Colin 27 atlas \citep{holmes1998enhancement}, the ICBM atlas \citep{mazziotta1995probabilistic,mazziotta2001probabilistic}, and the LONI LPBA40 atlas \citep{shattuck2008construction}. 

Computational atlas building using MRI is limited by the resolution and contrast that can be achieved with this imaging technique. The resolution barrier can be partly overcome with \emph{ex vivo} MRI, in which motion -- and hence time constraints -- are eliminated, enabling longer acquisition at ultra-high resolution ($\sim$\SI{100}{\micro\meter}), which in turns enable manual segmentation at a higher level of detail \citep{augustinack2005detection,yushkevich2009high,iglesias2015computational,saygin2017high}. However, not even the highest resolution achievable with \emph{ex vivo} MRI is sufficient to study microanatomy. Moreover, and despite recent advances in pulse sequences, MRI does not generate visible contrast at the boundaries of many neighbouring brain structures, the way that histological staining does.

For these reasons, recent studies building computational brain atlases are using stacks of digitised histological sections, which enable more accurate manual segmentations, to build atlases at a superior level of detail. Examples include the work by \citet{chakravarty2006creation} on the thalamus and basal ganglia; by \citet{krauth2010mean} on the thalamus; by \citet{adler2014histology,adler2016probabilistic} on the hippocampus; our recent work on the thalamus \citep{iglesias2017thalamus}, and the recently published atlas from the Allen Institute \citep{ding2016comprehensive}\footnote{\url{http://atlas.brain-map.org/atlas?atlas=265297126}}.

\subsection{Related work on 3D histology reconstruction}

The main drawback of building atlases with histology is  the fact that the 3D structure of the tissue is lost in the processing. Sectioning and mounting introduce large nonlinear distortions in the tissue structure, including artefacts such as folds and tears. In order to recover the 3D shape, image registration algorithms can be used to estimate the spatial correspondences between the different sections. This problem is commonly known as ``histology reconstruction''. 

The simplest approach to histology reconstruction is to sequentially align sections in the stack to their neighbours using a linear registration method. There is a wide literature on the topic, not only for histological sections but also for autoradiographs. Most of these methods use robust registration algorithms, e.g., based on edges \citep{hibbard1988objective,rangarajan1997robust}, block matching \citep{ourselin2001reconstructing} or point disparity \citep{zhao1993registration}. There are also nonlinear versions of serial registration methods (e.g., \citealt{arganda20103d,pitiot2006piecewise,chakravarty2006creation,schmitt2007image}), some of which introduce smoothness constraints to minimise the impact of sections that are heavily affected by artefacts and/or are poorly registered \citep{ju20063d,yushkevich20063d,cifor2011smoothness}.

The problem with serial alignment of sections is that, without any information on the original shape, methods are prone to straightening curved structures, a problem known as \emph{``z-shift''} or \emph{``banana effect''} (since the reconstruction of a sliced banana would be a cylinder). One way of overcoming this problem is the use of fiducial markers such as needles or rods (e.g., \citealt{humm2003stereotactic}); however, this approach has two disadvantages: the tissue may be damaged by the needles, and additional bias can be introduced in the registration if the sectioning plane is not perpendicular to the needles. 

Another way of combating the \emph{banana effect} is to use an external reference volume without geometric distortion. In an early study, \citet{kim1997mutual} used video frames to construct such reference, in the context of autoradiograph alignment. More recent works have used MRI scans (e.g.,  \citealt{malandain2004fusion,dauguet2007three,yang2012mri,ebner2017volumetric}). The general idea is to iteratively update: 1. a rigid transform bringing the MRI to the space of the histological stack; and 2. a nonlinear transform per histological section, which registers it to the space of the corresponding (resampled) MRI plane. A potential advantage of using MRI as a reference frame for histology reconstruction is that one recovers in MRI space the manual delineations made on the histological sections, which can be desirable when building atlases \citep{adler2016probabilistic}.

Increased stability in histology reconstruction can be obtained by using a third, intermediate modality to assist the process. Such modality is typically a stack of blockface photographs, which are taken prior to sectioning and are thus spatially undistorted. Such photographs help bridge the spaces of the MRI (neither modality is distorted) and the histology (plane correspondences are known). An example of this approach is the BigBrain project \citep{amunts2013bigbrain}. 

Assuming that a good estimate of the rigid alignment between the MRI and the histological stack is available,  the main technical challenge of 3D histology reconstruction is the nonlinear 2D registration of a histological section with the corresponding (resampled) MRI plane. These images exhibit very different contrast properties, in addition to modality-specific artefacts, e.g., tears in histology, bias field in MRI. Therefore, generic information theory based registration metrics such as mutual information \citep{maes1997multimodality,wells1996multi,pluim2003mutual} yield unsatisfactory results. This is partly due to the fact that such approaches only capture statistical relationships between image intensities at the voxel level, disregarding geometric information.

\subsection{Related work on image synthesis for registration}

An alternative to mutual information for inter-modality registration is to use image synthesis. The premise is simple: if we need to register a floating image $F_A$ of modality $A$ to a reference image $R_B$ of modality $B$, and we have access to a dataset of spatially aligned pairs of images of the two modalities $\{A_i,B_i\}$, then we can: estimate a synthetic version of the floating image $F_B$ that resembles modality $B$; register $F_B$ to $R_B$ with an intra-modality registration algorithm; and apply the resulting deformation field to the original floating image $F_A$. In the context of brain MRI, we have shown in \citet{iglesias2013synthesizing} that such an approach, even with a simple synthesis model \citep{hertzmann2001image}, clearly outperforms registration based on mutual information. This result has been replicated in other studies (e.g., \citealt{roy2014mr}), and similar conclusions have been reached in the context of MRI segmentation \citep{roy2013magnetic} and classification \citep{van2015does}.

Medical image synthesis has gained popularity in the last few years due to the advent of hybrid PET-MR scanners, since synthesising a realistic CT scan from the corresponding MR enables accurate attenuation correction of the PET data \citep{burgos2014attenuation,huynh2016estimating}. Another popular application of CT synthesis from MRI is dose calculation in radiation therapy \citep{kim2015implementation,siversson2015mri}. Unfortunately, most of these synthesis algorithms are based on supervised machine learning techniques, which require aligned pairs of images from the two modalities -- which are very hard to obtain for histology and MRI.

A possible alternative to supervised synthesis is a weakly supervised paradigm, best represented by the recent deep learning method CycleGAN \citep{CycleGAN2017}. This algorithm uses two sets of (unpaired) images of the two modalities, to learn two mapping functions, from each modality to the other. CycleGAN enforces cycle consistency of the two mappings (i.e., that they approximately invert each other), while training two classifiers that discriminate between synthetic and real images of each modality in order to avoid overfitting. While this technique has been shown to produce realistic medical images \citep{chartsias2017adversarial,wolterink2017deep}, it has an important limitation in the context of histology-MRI registration: it is unable to exploit the pairing between the (nonlinearly misaligned) histology and MRI images. Another disadvantage of CycleGAN is that, since a database of cases is necessary to train the model, it cannot be applied to a single image pair, i.e., it cannot be used as a generic inter-modality registration tool.

\subsection{Contribution}

In this study, we propose a novel probabilistic model that \emph{simultaneously} solves for registration and synthesis directly on the target images, i.e., without any training data. The principle behind the method is that improved registration provides less noisy data for the synthesis, while more accurate synthesis leads to better registration. Our framework enables these two components to iteratively exploit the improvements in the estimates of the other, while considering the uncertainty in each other's parameters. Taking uncertainty into account is crucial: if one simply tries to iteratively optimise synthesis and registration while keeping the other fixed to a point estimate, both components are greatly affected by the noise introduced by the other. More specifically, misregistration leads to bad synthesis due to noisy training data, whereas accurate registration to an poorly synthesised image yields incorrect alignment.

If multiple image pairs are available, the framework exploits the complete database, by jointly considering the probabilistic registrations between the pairs. In addition, the synthesis algorithm effectively takes advantage of the spatial structure in the data, as opposed to mutual information based registration. Moreover, the probabilistic nature of the model also enables the seamless integration of manually placed landmarks, which inform both the registration (directly) and the synthesis (indirectly, by creating areas of high certainty in the registration). We present a variational expectation maximisation algorithm  (VEM, also known as variational Bayes) to solve the model with Bayesian inference, and illustrate the proposed approach through experiments on synthetic and real data.

The rest of this paper is organised as follows. In Section~\ref{sec:methods}, we describe the probabilistic model on which our algorithm relies (Section~\ref{sec:probFramework}), as well as an inference algorithm to compute the most likely solution within the proposed framework (Section~\ref{sec:inference}). In Section~\ref{sec:ExpAndRes}, we describe the MRI and histological data (Section~\ref{sec:data}) that we used in our experiments (Section~\ref{sec:expSetup}), as well as the results on real data and the Allen atlas (Section~\ref{sec:results}). Finally, Section~\ref{sec:discussion} concludes the paper.


\section{Methods}
\label{sec:methods}

\subsection{Probabilistic framework}
\label{sec:probFramework}

\begin{figure*}[!t]
\centering
\subfigure[]{
\includegraphics[width=.34\textwidth]{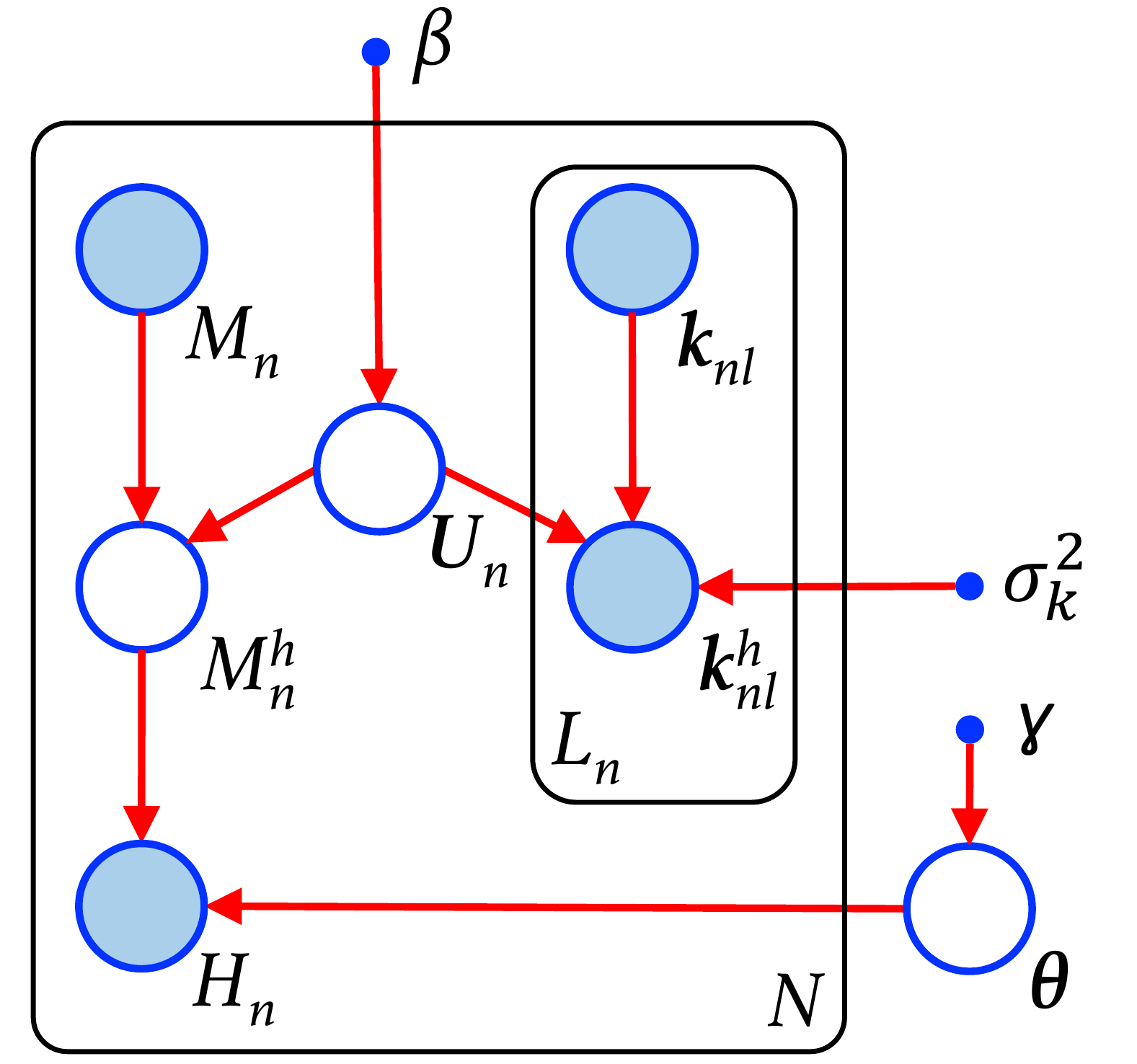}
}
~~
\subfigure[]{
  \begin{tabular}[b]{rl}
          $\bm{x}$ & Spatial coordinates  \\
          $\Omega_n$ & Image domain of $n^{th}$ image pair \\
          $N$ & Number of image pairs \\
          $M_n(\bm{x})$ & Intensities of $n^{th}$ MRI image \\
          $M_n^h(\bm{x})$ & Intensities of $n^{th}$ registered MRI image  \\
          $H_n(\bm{x})$ & Intensities of $n^{th}$ histological section  \\
          $\bm{U}_n(\bm{x})$ & Deformation field for $n^{th}$ image pair  \\
          $L_n$ & Number of available landmarks for $n^{th}$ image pair  \\
          $\bm{k}_{nl}$ & Spatial coordinates of $k^{th}$ landmark on $M_n$ \\
          $\bm{k}^h_{nl}$ & Spatial coordinates of $k^{th}$ landmark on $H_n$ \\
          $\theta$ & Parameters of image intensity transform (contrast synthesis) \\
          $\gamma$ & Hyperparameters of image intensity transform (contrast synthesis)  \\
          $\beta$ & Hyperparameters of deformation field \vspace{2pt} \\
          $\sigma_k^2$ & Variance of manual landmark placement \\
   \end{tabular}
}
\caption{(a) Graphical model of the proposed probabilistic framework. Circles represent random variables or parameters, arrows indicate dependencies between the variables, dots represent known (hyper)parameters, shaded variables are observed, and plates indicate replication. (b) Mathematical symbols corresponding to the model.}
\label{fig:graphicalModel}
\end{figure*}

The graphical model of our probabilistic framework and corresponding mathematical symbols are shown in Figure~\ref{fig:graphicalModel}.  For the sake of simplicity, we describe the framework from the perspective of the MRI to histology registration problem, though the method is general and can be applied to other inter-modality registration task -- in any number of dimensions.

Let $\{M_n\}_{n=1,\ldots,N}$ and $\{H_n\}_{n=1,\ldots,N}$ represent $N \geq 1$ MRI image slices and corresponding histological sections. We assume that each pair of images has been coarsely aligned with a 2D linear registration algorithm (e.g., using mutual information), and are hence defined over the same image domain $\Omega_n$. $M_n$ and $H_n$ are functions of the spatial coordinates $\bm{x}\in\Omega_n$, i.e., $M_n=M_n(\bm{x})$ and $H_n=H_n(\bm{x})$. In addition, let $K_n$ and $K^h_n$ represent two sets of $L_n$ corresponding landmarks, manually placed on the $n^{th}$ MRI image and histological section, respectively: $K_n=\{\bm{k}_{nl}\}_{l=1,\ldots,L_n}$ and $K^h_n=\{\bm{k}^h_{nl}\}_{l=1,\ldots,L_n}$, where $\bm{k}_{nl}$ and $\bm{k}^h_{nl}$ are 2D vectors with the spatial coordinates of the $l^{th}$ landmark on the $n^{th}$ image pair; for reasons that will be apparent in Section~\ref{sec:inference} below, we will assume that every $\bm{k}_{nl}$ coincides with an integer pixel coordinate. Finally, $M^h_n$ represents the $n^{th}$ MR image after applying a nonlinear deformation field $\bm{U}_n(\bm{x})$, which deterministically warps it to the space of the $n^{th}$ histological section $H_n$, i.e.,   
\begin{equation}
M^h_n(\bm{x}) = M_n(\bm{x} + \bm{U}_n(\bm{x})),
\label{eq:MnhModel}
\end{equation}
which in general requires interpolation of $M_n(\bm{x})$.

Each deformation field $\bm{U}_n$ is assumed to be an independent sample of a Markov Random Field (MRF) prior, with unary potentials penalising large shifts (their squared module), and binary potentials penalising the squared gradient magnitude:
\begin{equation}
p(\bm{U}_n) = \frac{1}{Z_n(\beta_1,\beta_2)}  \prod_{\bm{x}\in\Omega_n} e^{ - \beta_1 \| \bm{U}_n(\bm{x}) \|^2  - \beta_2 \sum_{\bm{x}'\in\mathcal{B}(\bm{x})}  \| \bm{U}_n(\bm{x}) - \bm{U}_n(\bm{x'}) \|^2 },
\label{eq:MRFprior}
\end{equation}
where $\beta_1 > 0$ and $\beta_2 > 0$ are the parameters of the MRF (which we group in $\beta=\{\beta_1,\beta_2\}$); $Z_n(\beta_1,\beta_2)$ is the partition function; and $\mathcal{B}(\bm{x})$ is the neighbourhood 
of the pixel located at $\bm{x}$. We note that this prior encodes a regularisation similar to that of the popular demons registration algorithm \citep{vercauteren2007non,cachier2003iconic}. Moreover, we also discretise the deformation field to a set of values (shifts) $\{\bm{\Delta}_s\}_{s=1,\ldots,S}$, i.e., $\bm{U}_n(\bm{x}) \in \{\bm{\Delta}_s\}$; we note that these shifts do not need to be integer (in pixels). While this choice of deformation model and regulariser does not guarantee the registration to be diffeomorphic (which might be desirable), it enables marginalisation over the deformation fields $\{\bm{U}_n\}$ -- and, as we will discuss in Section~\ref{sec:inference} below, a more sophisticated deformation model can be used to refine the final registration. 

Application of $\bm{U}_n$ to $M_n$ and $K_n$ yields not only a registered MRI image $M^h_n$ (Equation~\ref{eq:MnhModel}), but also a set of warped landmarks $K^h$. When modelling $K^h$, we need to account for the error made by the user when manually placing corresponding key-points in the MR images and the histological sections. We assume that these errors are independent and follow zero-mean, isotropic Gaussian distributions parametrised by their  covariances $\sigma_k^2 \bm{I}$ (where $\bm{I}$ is the 2$\times$2 identity matrix, and where $\sigma_k^2$ is expected to be quite small):
\begin{align}
p(K_n^h | K_n, \bm{U}_n, \sigma_k^2) = & \prod_{l=1}^{L_n} p(\bm{k}_{nl}^h | \bm{k}_{nl} - \bm{U}_n(\bm{k}_{nl}^h), \sigma_k^2) \nonumber\\
= & \prod_{l=1}^{L_n} \frac{1}{2\pi\sigma_k^2} \exp\left[ -\frac{1}{2\sigma_k^2} \| \bm{k}_{nl}^h - \bm{k}_{nl} + \bm{U}_n(\bm{k}_{nl}^h) \|^2 \right].
\label{eq:LMmodel}
\end{align}

Finally, to model the connection between the intensities of the histological sections $\{H_n\}$ and the registered MRI images $\{M_n^h$\}, we follow \citet{tu2008brain} and make the assumption that:
\begin{equation}
p(H_n|M_n^h,\theta) \propto p(M_n^h|H_n,\theta).  
\label{eq:cheat}
\end{equation}
This assumption is equivalent to adopting a discriminative approach to model the contrast synthesis. While this discriminative component breaks the generative nature of the framework, it also enables the modelling of much more complex relationships between the intensities of the two modalities, including spatial and geometric information about the pixels. Such spatial patterns cannot be captured by, e.g., mutual information, which only models statistical relationships between intensities (e.g., a random shuffling of pixels does not affect the metric). Here we use a regression forest \citep{breiman2001random}, though any other discriminative regression could have been utilised.  We assume conditional independence of the pixels in the prediction: the forest produces a Gaussian distribution for each pixel $\bm{x}$ separately, parametrised by $\mu_{n\bm{x}}$ and $\sigma^2_{n\bm{x}}$. Moreover, we place a (conjugate) Inverse Gamma prior on the variances $\sigma^2_{n\bm{x}}$, with hyperparameters $a$ and $b$:
\begin{equation}
p(\sigma^2_{n\bm{x}} | a,b) =  \frac{b^a}{\Gamma(a)}(\sigma^2_{n\bm{x}})^{-a-1}\exp(-b/\sigma^2_{n\bm{x}}).
\label{eq:IGprior}
\end{equation}
We use $\theta$ to represent the set of forest parameters, which groups the selected features, split values, tree structure and the prediction at each leaf node. The set of corresponding hyperparameters are grouped in $\gamma$, which includes the parameters of the Gamma prior $\{a,b\}$, the number of trees, maximum depth, and minimum number of samples in leaf nodes. The intensity model is hence:
\begin{align}
p(M_n^h & | H_n, \theta) = \prod_{\bm{x}\in \Omega_n}  p \left( M^h_n(\bm{x}) | H_n(\mathcal{W}(\bm{x})),\theta \right) \nonumber\\
= & \prod_{\bm{x}\in \Omega_n} \mathcal{N}\left( M^h_n(\bm{x}); \mu_{n\bm{x}}(H_n(\mathcal{W}(\bm{x})),\theta),\sigma^2_{n\bm{x}}(H_n(\mathcal{W}(\bm{x})),\theta) \right), \nonumber
\end{align}
where $\mathcal{W}(\bm{x})$ is a spatial window centred at $\bm{x}$, and $\mathcal{N}$ represents the Gaussian distribution. Given the deterministic deformation model (Equation~\ref{eq:MnhModel}), and the assumption in Equation~\ref{eq:cheat}, we finally obtain the likelihood term: 
\begin{align}
p(H_n&  | M_n , \bm{U}_n, \theta ) = \prod_{\bm{x}\in \Omega_n}  p \left( M_n(\bm{x}+\bm{U}(\bm{x})) | H_n(\mathcal{W}(\bm{x})),\theta \right) \nonumber\\
= & \prod_{\bm{x}\in \Omega_n} \mathcal{N}\left( M_n(\bm{x}+\bm{U}(\bm{x})); \mu_{n\bm{x}}(H_n,\theta),\sigma^2_{n\bm{x}}(H_n(\mathcal{W}(\bm{x})),\theta) \right).\label{eq:IntensityModel}
\end{align}

We emphasise that, despite breaking the generative nature of the model, the assumption in Equation~\ref{eq:cheat} still leads to a valid objective function when performing Bayesian inference. This cost function can be optimised with standard inference techniques, as explained in Section~\ref{sec:inference} below.

\subsection{Inference}
\label{sec:inference}

The final goal is to estimate the registrations $\{\bm{U}_n\}$. To do so, we use Bayesian inference to ``invert'' the probabilistic model described in  Section~\ref{sec:probFramework} above. If we group all the observed variables into the set $O=\{ \{M_n\},\{H_n\},\{K_n\},\{K^h_n\},\beta,\gamma,\sigma^2_k\}$, the problem is to maximise the posterior probability of the registrations given the available information, i.e., $p(\{\bm{U}_n\}|O)$. Computing such probability distribution requires marginalizing over the intensity model, which leads to an intractable integral:
\begin{align}
\{\hat{\bm{U}_n}\} & = \argmax_{\{\bm{U}_n\}}  p( \{\bm{U}_n\} | O) \nonumber \\ 
& = \argmax_{\{\bm{U}_n\}} \int_{\theta} p( \{\bm{U}_n\} | \theta, O) p(\theta | O) d\theta. \label{eq:intractable}
\end{align}
To solve this problem, we make the standard approximation that the posterior distribution of the parameters $\theta$ given the observed data $O$ is strongly peaked around its mode $\hat\theta$, i.e., we use its point estimate. The intractable integral in Equation~\ref{eq:intractable} then becomes:
\begin{equation}
\{\hat{\bm{U}_n}\} = \argmax_{\{\bm{U}_n\}} p( \{\bm{U}_n\} | \hat\theta, O),
 \label{eq:finalRegistrations}
\end{equation}
where the point estimate is given by:
\begin{equation}
\hat\theta = \argmax_{\theta} p(\theta | O).
\label{eq:pointEstimate}
\end{equation}
In this section, we first describe a VEM algorithm to obtain the point estimate of $\theta$ using Equation~\ref{eq:pointEstimate} (Section~\ref{sec:pointEstimate}), and then address the computation of the final registrations with Equation~\ref{eq:finalRegistrations} (Section~\ref{sec:finalRegistrations}).

\subsubsection{Computation of point estimate $\hat\theta$}
\label{sec:pointEstimate}

Applying Bayes's rule on Equation~\ref{eq:pointEstimate}, and taking the logarithm, we obtain the following objective function:
\begin{align}
\hat\theta = &  \argmax_{\theta} p(\theta | \{M_n\},\{H_n\},\{K_n\},\{K^h_n\},\beta,\gamma,\sigma^2_k) \nonumber \\
 = & \argmax_{\theta} \log p(\{K^h_n\}, \{H_n\} | \theta, \{M_n\},\{K_n\},\beta,\gamma,\sigma^2_k) + \log p(\theta | \gamma). \label{eq:objectiveFun}
\end{align}
Exact maximisation of Equation~\ref{eq:objectiveFun} would require marginalizing over the deformation fields $\{\bm{U}_n\}$, which leads (once again) to an intractable integral due to the pairwise terms of the MRF prior (Equation~\ref{eq:MRFprior}). Instead, we use a variational technique (VEM) for approximate inference. Since the Kullback-Leibler (KL) divergence is by definition non-negative, the objective function in Equation~\ref{eq:objectiveFun} is bounded from below by:
\begin{small}
\begin{align}
J[q&(\{\bm{U}_n\}),\theta] = \log p(\{K^h_n\}, \{H_n\} | \theta, \{M_n\},\{K_n\},\beta,\gamma,\sigma^2_k\}) + \log p(\theta | \gamma) \nonumber \\
  & - KL [q(\{\bm{U}_n\}) \| p(\{\bm{U}_n\} | \{K^h_n\}, \{H_n\}, \theta, \{M_n\},\{K_n\},\beta,\gamma,\sigma^2_k\}) \label{eq:targetE}\\
   = & \eta[q] + \sum_{\{\bm{U}_n\}} q(\{\bm{U}_n\}) \log p(\{\bm{U}_n\} , \{K^h_n\}, \{H_n\}  | \theta, \{M_n\},\{K_n\},\beta,\gamma,\sigma^2_k\}) \nonumber \\
  & + \log p(\theta | \gamma). \label{eq:targetM}
\end{align}
\end{small}
The bound $J[q(\{\bm{U}_n\}),\theta]$ is the negative of the so-called free energy: $\eta$ represents the entropy of a random variable; and $q(\{\bm{U}_n\})$ is a distribution over $\{\bm{U}_n\}$ which approximates the posterior $p(\{\bm{U}_n\} | \{K^h_n\}, \{H_n\}, \theta, \{M_n\},\{K_n\},\beta,\gamma,\sigma^2_k\})$, while being restricted to have a simpler form. The standard mean field approximation \citep{parisi1988statistical} assumes that $q$ factorises over voxels for each field $\bm{U}_n$:
$$
q(\{\bm{U}_n\}) = \prod_{n=1}^N \prod_{\bm{x}\in\Omega_n} q_{n\bm{x}}(\bm{U}_n(\bm{x})),
$$
where $q_{n\bm{x}}$ is a discrete distribution over shifts at pixel $\bm{x}$ of image $n$, such that $q_{n\bm{x}}(\bm{\Delta}_s)\geq 0$, $\sum_{s=1}^S q_{n\bm{x}}(\bm{\Delta_s}) = 1$, $\forall n, \bm{x}$.

Rather than the original objective function (Equation~\ref{eq:objectiveFun}), VEM maximises the lower bound $J$, by alternately optimizing with respect to $q$ (E-step) and $\theta$ (M-step) in a coordinate ascent scheme. We summarise these two steps below.

\begin{figure*}
\centering
\includegraphics[width=0.9\textwidth]{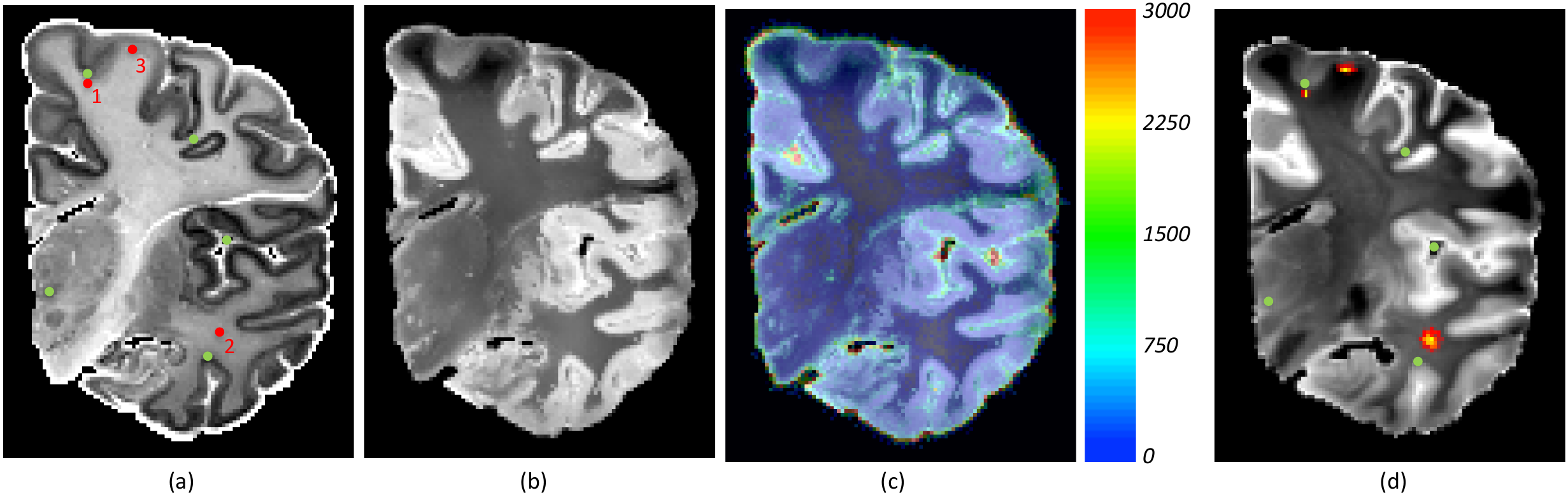}
\caption{Illustration of the VEM algorithm: (a) Histological section from the Allen atlas. The green dots represent manually placed landmarks. (b) Mean of synthesised MRI slice, after 5 iterations of the VEM algorithm. (c) Variance of synthesised MRI slice, overlaid on the mean. (d) Corresponding real MRI slice. The green dots represent the manually placed landmarks, corresponding to the ones in (a). The heat maps represent the approximate posterior distributions of shifts ($q_{n\bm{x}}$) corresponding to the red dots in (a).}
\label{fig:exampleImages}
\end{figure*}

\paragraph{E-step} To optimise the lower bound with respect to $q$, it is convenient to work with Equation~\ref{eq:targetE}. Since the first two terms are independent of $q$, one can minimise the KL divergence between $q$ and the posterior distribution of $\{\bm{U}_n\}$:
\begin{small}
\begin{align}
\argmin_q  & \sum_{n=1}^N \sum_{\bm{x}\in\Omega_n} \sum_{s=1}^S q_{n\bm{x}}(\bm{\Delta}_s) \log \frac{q_{n\bm{x}}(\bm{\Delta}_s)}{p \left( M_n(\bm{x}+\bm{\Delta}_s) | H_n(\mathcal{W}(\bm{x})),\theta \right) } \nonumber \\
 - & \sum_{n=1}^N \sum_{\bm{x}\in\Omega_n} \sum_{s=1}^S q_{n\bm{x}}(\bm{\Delta}_s) \log \left({ e^{-\beta_1 \|\bm{\Delta}_s\|^2}  \prod_{l=1}^{L_n}  p \left(\bm{k}_{nl}^h | \bm{k}_{nl} - \bm{\Delta}_s, \sigma_k^2 \right)^{\delta(\bm{k}_{nl}=\bm{x})}}\right) \nonumber \\
- & \beta_2 \sum_{n=1}^N \sum_{\bm{x}\in\Omega_n} \mathbb{E}_{q_{n\bm{x}}} \left[  \sum_{\bm{x}'\in\mathcal{B}(\bm{x})} \sum_{s=1}^S \|\bm{U}_{n}(\bm{x}) - \bm{\Delta}_{s}\|^2  q_{n\bm{x}'} (\bm{U}_{n}(\bm{x})) \right], \nonumber 
\end{align}
\end{small}
where $\mathbb{E}$ is the expected value. Building the Lagrangian (to ensure that $q$ stays in the probability simplex) and setting derivatives to zero, we obtain:
\begin{align}
q_{n\bm{x}}(\bm{\Delta}_s) & \propto 
p \left( M_n(\bm{x}+\bm{\Delta}_s) | H_n(\mathcal{W}(\bm{x})),\theta \right) e^{-\beta_1 \|\bm{\Delta}_s\|^2} \nonumber \\
& \times \prod_{l=1}^{L_n}  p\left(\bm{k}_{nl}^h | \bm{k}_{nl} - \bm{\Delta}_s, \sigma_k^2\right)^{\delta(\bm{k}_{nl}=\bm{x})} \nonumber \\
& \times \exp\left(\beta_2  \sum_{\bm{x}'\in\mathcal{B}(\bm{x})} \sum_{s'=1}^S \| \bm{\Delta}_s - \bm{\Delta}_{s'} \|^2  q_{n\bm{x}'} (\bm{\Delta}_{s'})   \right).
\label{eq:fixedPointIts}
\end{align}
This equation has no closed-form solution, but can be solved with fixed point iterations, one image pair at the time -- since there is no interdependence in $n$. We note that the effect of the landmarks is not local; in addition to creating a very sharp $q_{n\bm{x}}$ around pixel at hand, the variational algorithm also creates a high confidence region around $\bm{x}$, by encouraging neighbouring pixels to have similar shifts. This is illustrated in Figure~\ref{fig:exampleImages}(a,d). The spatial location marked by red dot number 1 is right below a manually placed landmark in the histological section, and the distribution $q_{n\bm{x}}$ is hence strongly peaked at a location right below the corresponding landmark in the MRI slice. Red dot number 2, on the contrary, is located in the middle of the cerebral white matter, where there is little contrast to guide the registration, so $q_{n\bm{x}}$ is much more more spread and isotropic. Red dot number 3 lies in the white matter right under the cortex, so its distribution is elongated and parallel to the white matter surface.

\paragraph{M-step} When optimizing $J$ with respect to $\theta$, it is more convenient to work with Equation~\ref{eq:targetM} -- since the term $\eta[q]$ can be neglected. Applying the chain rule of probability, and leaving aside terms independent of $\theta$, we obtain:
\begin{align}
\argmax_\theta & \sum_{\{\bm{U}_n\}} q(\{\bm{U}_n\}) \log p(\{H_n\} | \{\bm{U}_n\}, \{M_n\}, \theta) + \log p(\theta | \gamma)  \nonumber \\
= \argmax_\theta & \sum_{n=1}^N \sum_{\bm{x}\in\Omega_n} \sum_{s=1}^S  q_{n\bm{x}}(\bm{\Delta}_s) \log p(M_n(\bm{x}+\bm{\Delta}_s) | H_n(\mathcal{W}(\bm{x})), \theta) \nonumber \\
 & +  \log p(\theta | \gamma).  \label{eq:MstepObjective}
\end{align}
Maximisation of Equation~\ref{eq:MstepObjective} amounts to training the regressor, such that each input image patch $H_n(\mathcal{W}(\bm{x}))$ is considered $S$ times, each with an output intensity corresponding to a differently shifted pixel location $M_n(\bm{x}+\bm{\Delta}_s)$, and with weight $q_{n\bm{x}}(\bm{\Delta}_s)$. In practice,  and since injection of randomness is a crucial aspect of the training process of random forests, we found it beneficial to consider each patch $H_n(\mathcal{W}(\bm{x}))$ only once in each tree, with a shift $\bm{\Delta}_s$ sampled from the corresponding distribution $q_{n\bm{x}}(\bm{\Delta})$ -- fed to the tree with weight $1$.  

The injection of additional randomness through sampling of $\bm{\Delta}$ no only greatly increases the robustness of the regressor against misregistration, but also decreases the computational cost of training -- since only a single shift is considered per pixel. We also note that this sampling strategy still yields a valid stochastic optimiser for Equation~\ref{eq:MstepObjective}, since $q_{n\bm{x}}$ is a discrete probability distribution over shifts. Such stochastic procedure (as well as other sources of randomness in the forest training algorithm) makes the maximisation of Equation~\ref{eq:MstepObjective} only approximate; this means that the coordinate ascent algorithm to maximise the lower bound $J$ of the objective function is no longer guaranteed to converge. In practice, however, the VEM algorithm typically converges after $\sim$5 iterations. 

Combined with the conjugate prior on the variance $p(\theta | \gamma)$, the joint prediction of the forest is finally given by:
\begin{align}
\mu_{n\bm{x}} = & \frac{1}{T} \sum_{t=1}^T g_t[H_n(\mathcal{W}(\bm{x})); \theta] \nonumber \\
\sigma^2_{n\bm{x}} =& \frac{2b + \sum_{t=1}^T \left( g_t[H_n(\mathcal{W}(\bm{x})); \theta] - \mu_{n\bm{x}} \right)^2}{2a + T}, \label{eq:forestPreds}
\end{align}
where $g_t$ is the guess made by tree $t$;  $T$ is the total number of trees in the forest; and where we have dropped the dependency of $\mu_{n\bm{x}}$ and $\sigma_{n\bm{x}}$ on $\{H_n,\hat\theta\}$ for simplicity. A sample output of the forest is shown in Figure~\ref{fig:exampleImages}(b,c). Areas of higher uncertainty, which will be downweighted in the registration, include the horizontal crack on the histological image and cerebrospinal fluid regions; the latter may appear bright or dark, depending on whether they are filled with paraformaldehyde, air or Fomblin (further details on these data can be found in Section~\ref{sec:data}).

\subsubsection{Computation of optimal deformation fields $\{\hat{\bm{U}}_n\}$}
\label{sec:finalRegistrations}

Once the point estimate $\hat\theta$ (i.e., the optimal regression forest for synthesis) has been computed, one can obtain the optimal registrations by maximizing $p(\{\bm{U}_n\} | \hat\theta,\{M_n\},\{H_n\},\{K_n\},\{K^h_n\},\beta,\sigma^2_k)$. Using Bayes's rule and taking the logarithm, we obtain: 
\begin{align}
\argmax_{\{\bm{U}_n\}} & \sum_{n=1}^N \sum_{\bm{x}\in\Omega_n} 
\log p \left( M_n(\bm{x}+\bm{U}_n(\bm{x})) | H_n(\mathcal{W}(\bm{x})),\hat\theta \right) \nonumber \\
+ & \sum_{n=1}^N \sum_{l=1}^{L_n} p(\bm{k}^h_{nl} | \bm{k}_{nl} - \bm{U}_n(\bm{k}^h_{nl}))  -\beta_1  \sum_{n=1}^N \sum_{\bm{x}\in\Omega_n}  \|\bm{U}_n(\bm{x}))\|^2  \nonumber \\
- & \beta_2  \sum_{n=1}^N  \sum_{\bm{x}\in\Omega_n} \sum_{\bm{x'}\in\mathcal{B}(\bm{x})}  \| \bm{U}_n(\bm{x}) - \bm{U}_n(\bm{x'}) \|^2 ,
\label{eq:optDef}
\end{align}
which can be solved one image pair $n$ at the time. Substituting the Gaussian likelihoods into Equation~\ref{eq:optDef}, switching signs and disregarding terms independent of $\bm{U}_n$ yields, for each image pair, the following cost function for the registration is obtained:
\begin{align}
\hat{\bm{U}}_n & =  
\argmin_{\bm{U}_n} \underbrace{\sum_{\bm{x}\in\Omega_n}  \frac{\left[ M_n(\bm{x}+\bm{U}_n(\bm{x})) - \hat\mu_{n\bm{x}} \right]^2}{2\hat\sigma^2_{n\bm{x}} }}_\text{Image term} \nonumber \\
 & + \underbrace{\frac{1}{2\sigma_k^2} \sum_{l=1}^{N_l}  \| \bm{k}^h_{nl} - \bm{k}_{nl} + \bm{U}_n(\bm{k}^h_{nl})\|^2}_\text{Landmark term}  \nonumber \\
&  + \underbrace{\beta_1  \sum_{\bm{x}\in\Omega_n}  \|\bm{U}_n(\bm{x}))\|^2 +  \beta_2 \sum_{n=1}^N  \sum_{\bm{x}\in\Omega_n} \sum_{\bm{x'}\in\mathcal{B}(\bm{x})}  \| \bm{U}_n(\bm{x}) - \bm{U}_n(\bm{x'}) \|^2}_\text{Regularisation}.  \label{eq:costReg}
\end{align}
Thanks to the discrete nature of $\bm{U}_n$, a local minimum of the cost function in Equation~\ref{eq:costReg} can be efficiently found with algorithms based on graph cuts \citep{ahuja1993network}, such as \citet{boykov2001fast}.   

We note that the result does not need to be diffeomorphic or invertible, which might be a desirable feature of the registration. This is due to the properties of the deformation model, which was chosen due to the fact that it easily enables marginalisation over the deformation fields with variational techniques. In practice, we have found that, once the optimal (probabilistic) synthesis has been computed, we can obtain smoother and more accurate solutions by using more sophisticated deformation models and priors. More specifically, we implemented the image and landmark terms of Equation~\ref{eq:costReg} in NiftyReg \citep{modat2010fast}, which is a fast, powerful registration package, instantly getting access to its advanced, efficiently implemented deformation models, regularisers and optimisers. NiftyReg parametrises the deformation field with a grid of control points combined with cubic B-Splines \citep{rueckert1999nonrigid}. If $\bm{\Psi}_n$ represents the vector of parameters of the spatial transform $\bm{x}' = \bm{V}(\bm{x};\bm{\Psi}_n)$ for image pair $n$, we optimise:
\begin{align} 
\hat{\bm{\Psi}}_n    =  
\argmin_{\bm{\Psi}_n}  \quad & \alpha  \sum_{\bm{x}\in\Omega_n}  \frac{\left[ M_n(\bm{V}(\bm{x};\bm{\Psi}_n)) - \hat\mu_{n\bm{x}} \right]^2}{2\hat\sigma^2_{n\bm{x}} }  \nonumber \\
 & + \frac{1}{2\sigma_k^2} \sum_{l=1}^{N_l}  \| \bm{V}(\bm{k}^h_{nl};\bm{\Psi}_n)    - \bm{k}_{nl}  \|^2 \nonumber \\
 & + \beta_b E_b(\bm{\Psi}_n) + \beta_l E_l(\bm{\Psi}_n) + \beta_j E_j(\bm{\Psi}_n),  \label{eq:costRegNiftyReg}
\end{align}
where $E_b(\bm{\Psi}_n)$ is the bending  energy of the transform parametrised by $\bm{\Psi}_n$; $E_l(\bm{\Psi}_n)$  is the sum of squares of the symmetric part of the Jacobian after filtering out rotation (penalises stretching and shearing);  $E_j$ is the Jacobian energy (given by its log-determinant); $\beta_b>0$, $\beta_l>0$, $\beta_j>0$ are the corresponding weights; and $\alpha>0$ is a constant that scales the contribution of the image term, such that it is approximately bounded by 1: $\alpha^{-1}= 9 |\Omega_n| / 2 $, i.e., a value of 1 is achieved if all pixels are three standard deviations away from the predicted mean.

Note that this choice for the final model also enables comparison with mutual information as implemented in NiftyReg, which minimises:
\begin{align} 
\hat{\bm{\Psi}}_n^{MI}   =  
\argmin_{\bm{U}_n} \; & - \text{MI}[M_n(V(\bm{x};\bm{\Psi}_n)),H_n(\bm{x})]   \nonumber \\
 & + \frac{1}{2\sigma_k^2} \sum_{l=1}^{N_l}  \| \bm{V}(\bm{k}^h_{nl};\bm{\Psi}_n)    - \bm{k}_{nl}  \|^2 \nonumber \\
 & + \beta_b E_b(\bm{\Psi}_n) + \beta_l E_l(\bm{\Psi}_n) + \beta_j E_j(\bm{\Psi}_n),  \label{eq:costRegMIinNiftyReg}
\end{align}
where MI represents the mutual information. We note that finding the value of $\alpha$ that matches the importances of the data terms in Equations \ref{eq:costRegNiftyReg} and \ref{eq:costRegMIinNiftyReg} is a non-trivial task; however, our choice of $\alpha$ defined above places the data terms in approximately the same range of values.

\subsection{Summary of the algorithm and implementation details}
\label{sec:summaryAlgorithm}

The presented method is summarised in Algorithm~\ref{alg:method}. The approximate posteriors $q_{n\bm{x}}(\bm{\Delta})$ are initialised to $1/S$, evenly spreading the probability mass across all possible shifts (i.e., maximum uncertainty in the registration). Given $q_{n\bm{x}}$, Equation~\ref{eq:MstepObjective} is used to initialise the forest parameters $\theta$. At that point, the VEM algorithm alternates between the E and M steps until convergence is reached.  Convergence would ideally be assessed with $\theta$ but, since these parameters vary greatly from one iteration to the next due to the randomness injected in training, we use the predicted means and variances instead ($\mu_{n\bm{x}},\sigma^2_{n\bm{x}}$).

\begin{algorithm}
 \caption{Simultaneous synthesis and registration}
 \label{alg:method}
 \begin{algorithmic}
 \REQUIRE $\{M_n\}_{n=1,\ldots,N}$, $\{H_n\}_{n=1,\ldots,N}$, $K_n$, $K_n^h$
 \ENSURE $\hat\theta$, $\{\hat{\bm{U}}_n\}$
 \STATE $q_{n\bm{x}}(\bm{\Delta}) \leftarrow 1/S, \forall n, \bm{x}$
 \STATE Initialise $\theta$ with Eq.~\ref{eq:MstepObjective} (random forest training)
 \WHILE{$\mu_{n\bm{x}},\sigma^2_{n\bm{x}}$ change}
 \STATE \emph{E-step:}
 \FOR{$n=1$ to $n=N$}
 \STATE Compute $\mu_{n\bm{x}},\sigma^2_{n\bm{x}}, \forall \bm{x}\in\Omega_n$ with Eq.~\ref{eq:forestPreds}
 \WHILE{$q_{n\bm{x}}$ changes}
 \STATE Fixed point iteration of $q_{n\bm{x}}$ (Eq.~\ref{eq:fixedPointIts})
 \ENDWHILE
 \ENDFOR
 \STATE \emph{M-step:}
 \STATE Update $\theta$ with Eq.~\ref{eq:MstepObjective} (random forest retraining)
 \ENDWHILE
 \STATE $\hat\theta \leftarrow \theta$
  \FOR{$n=1$ to $n=N$}
 \STATE Compute final $\mu_{n\bm{x}},\sigma^2_{n\bm{x}}, \forall \bm{x}\in\Omega_n$ with Eq.~\ref{eq:forestPreds}
 \STATE Compute $\hat{\bm{U}}_n$ with Eq.~\ref{eq:costReg} or Eq.~\ref{eq:costRegNiftyReg}
 \ENDFOR
  \end{algorithmic}
 \end{algorithm}

In the E-step, each image pair can be considered independently. First, the histological section is pushed through the forest to generate a prediction for the (registered) MR image, including a mean and a standard deviation for each pixel (Equation~\ref{eq:forestPreds}). Then, fixed point iterations of Equation~\ref{eq:fixedPointIts} are run until convergence of $q_{n\bm{x}}, \forall \bm{x}\in\Omega_n$. In the M-step, the approximate posteriors $q$ of all images are used together to retrain the random forest with  Equation~\ref{eq:MstepObjective}. When the algorithm has converged, the final predictions (mean, variance) can be generated for each voxel, and the final registrations can be computed with Equation~\ref{eq:costReg}, or with NiftyReg (see details below).

Injection of randomness is a crucial aspect of random forests, as it increases their generalization ability \citep{criminisi2011decision}. Here we used bagging \citep{breiman1996bagging} at both the image and pixel levels, and used random subsets of features when splitting data at the internal nodes of the trees. An additional random component in the stochastic optimization is the sampling of shifts $\bm{\Delta}$ to make the model robust against misregistration (see Section~\ref{sec:pointEstimate}). While all these random elements have beneficial effects, these come at the expense of giving up the theoretical guarantees on the convergence of the VEM algorithm -- though this was never found to be a problem in practice, as explained in Section~\ref{sec:pointEstimate} above.    
 
For the final registration, we used the default regularisation scheme in NiftyReg, which is a weighted combination of the bending energy (second derivative) and the sum of squares of the symmetric part of the Jacobian. We note that  NiftyReg uses $\beta_j  = 0$ by default; while using $\beta_j  > 0$ guarantees that the output is diffeomorphic, the other two regularisation terms ($E_b, E_l$) ensure in practice that the deformation field is well behaved.


\section{Experiments and results}
\label{sec:ExpAndRes}

\subsection{Data}
\label{sec:data}

We used two datasets to validate the proposed technique; one synthetic, and one real. The synthetic data, which consists only of MRI images, enables quantitative comparison of the estimated deformations with the ground truth fields that were used to generate them. The real dataset, on the other hand, enables qualitative evaluation in a real histology-MRI registration problem. 

\subsubsection{Synthetic MRI dataset}

Since obtaining  MRI and histological data with perfect spatial alignment is very difficult, we used a synthetic dataset based solely on MRI to quantitatively validate the proposed approach. These synthetic data were generated from 676 (real) pairs of T1- and T2-weighted scans from the publicly available ADNI dataset. The resolution of the T1 scans was approximately 1 mm isotropic; the ADNI project spans multiple sites, different scanners were used to acquire the images; further details on the acquisition can be found
 at \url{http://www.adni-info.org}. The T2 scans correspond to an acquisition designed to study the hippocampus, and consist of 25-30 coronal images at 0.4$\times$0.4 mm resolution, with slice thickness of 2 mm. These images cover a slab of tissue containing the hippocampi, which is manually oriented by the operator to be approximately orthogonal to the major axes of the hippocampi. Once more, further details on the acquisition at different sites can be found at the ADNI website. 
 
The T1 scans were preprocessed with FreeSurfer \citep{fischl2012freesurfer} in order to obtain skull-stripped, bias-field corrected images with a corresponding segmentation of brain structures \citep{fischl2002whole}. We simplified this segmentation to three tissue types (gray matter, white matter, cerebrospinal fluid)   and a generic background label. The processed T1 was rigidly registered to the corresponding T2 scan with mutual information, as implemented in NiftyReg \citep{modat2014global}. The registration was also used to propagate the brain mask and automated segmentation; the former was used to skull-strip the T2, and the latter for bias field correction using the technique described in \citet{van1999automated}. Note that we deform the T1 to the T2 -- despite its lower resolution -- because of its more isotropic voxel size. 

From these pairs of preprocessed 3D scans, we generated a dataset of 1000 pairs of 2D images. To create each image pair, we followed these steps: 
1. Randomly select one pair of 3D scans;
2. In the preprocessed T2 scan, randomly select a (coronal) slice, other than the first and the last, which sometimes display artefacts; 
3. Downsample the T2 slice to 1 $\times$ 1 mm resolution, for consistency with the resolution of the  T1 scans; 
4. Reslice the (preprocessed) T1 scan to obtain the 2D image corresponding to the downsampled T2 slice; 
5. Sample a random diffeomorphic deformation field (details below) in the space of the 2D slice; 
6. Combine the deformation field with a random similarity transform, including rotation, scaling and translation;
7. Deform the T2 scan with the composed field (linear + nonlinear).  
8. Rescale intensities to [0,255] and discretise with 8-bit precision.

\begin{figure*}
\centering
\includegraphics[width=0.8\textwidth]{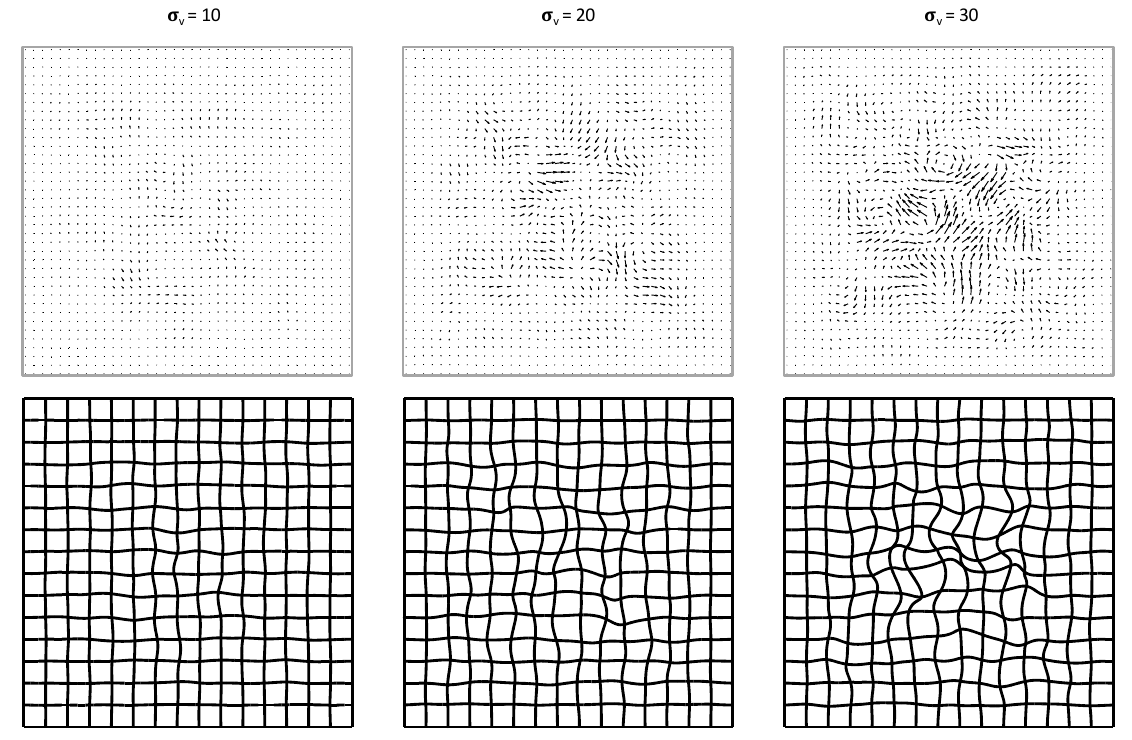}
\caption{Synthetic velocity (top row) and corresponding deformation fields (bottom row) generated with three different levels of noise $\sigma_v$.}
\label{fig:fields}
\end{figure*}

To generate synthetic fields without biasing the evaluation, we used a deformation model different from that used by NiftyReg (i.e., a grid of control points and cubic B-Splines). More specifically, we created diffeormorphic deformations as follows. First, we generated random velocity fields by independently sampling bivariate Gaussian noise at each spatial location (no x-y correlation) with different levels of variance;  smoothing them with a Gaussian filter; and multiplying them by a window function in order to prevent deformations close to the boundaries; we used $\exp[0.01 D(\bm{x})]$, where $D(\bm{x})$ is the distance to the boundary of the image in mm. Then, these velocity fields were integrated over unit time using a scaling and squaring approach \citep{moler2003nineteen,arsigny2006log} to generate the deformation fields. Sample velocity and deformation fields generated with different levels of noise are shown in Figure~\ref{fig:fields}.

Finally, we synthetically generated spatially spread landmarks using the following procedure. First, we calculated the response to a Harris corner detector \citep{harris1988combined}. Then, we iteratively selected the pixel with the highest response $\bm{x}_{max}$, and multiplied the Harris response by a complementary Gaussian function centred at $\bm{x}_{max}$, i.e.,
$$
f(\bm{x}) = 1-\exp[-0.5 \|\bm{x}-\bm{x}_{max}\|^2 /\sigma^2],
$$
with standard deviation $\sigma$ equal to 1/10 of the image dimensions. We then applied the deformation field to the landmarks, and corrupted the output locations with Gaussian noise of variance $\sigma_k^2$.

The ADNI was launched in 2003 by the National Institute on Aging, the National Institute of Biomedical Imaging and Bioengineering, the Food and Drug Administration, private pharmaceutical companies and non-profit organisations, as a \$60 million, 5-year public-private partnership. The main goal of ADNI is to test whether MRI, positron emission tomography (PET), other biological markers, and clinical and neuropsychological assessment can be combined to analyse the progression of MCI and early AD. Markers of early AD progression can aid researchers and clinicians to develop new treatments and monitor their effectiveness, as well as decrease the time and cost of clinical trials. The Principal Investigator of this initiative is Michael W. Weiner, MD, VA Medical Center and University of California - San Francisco. ADNI is a joint effort by co-investigators from industry and academia. Subjects have been recruited from over 50 sites across the U.S. and Canada. The initial goal of ADNI was to recruit 800 subjects but ADNI has been followed by ADNI-GO and ADNI-2. These three protocols have recruited over 1,500 adults (ages 55-90) to participate in the study, consisting of cognitively normal older individuals, people with early or late MCI, and people with early AD. The follow up duration of each group is specified in the corresponding protocols for ADNI-1, ADNI-2 and ADNI-GO. Subjects originally recruited for ADNI-1 and ADNI-GO had the option to be followed in ADNI-2.

\subsubsection{Real data}

To qualitatively evaluate our algorithm on real data, we used the Allen atlas, which is based on the left hemisphere of a 34-year-old donor. The histology of the atlas includes 106  Nissl-stained sections of the whole hemisphere in coronal plane, with manual segmentations of 862 brain structures. Due to the challenges associated with sectioning and mounting thin sections from complete hemispheres, artefacts such as cracks are present in the sections, which make registration difficult (see for intance Figure~\ref{fig:exampleImages}a). The sections are \SI{50}{\micro\meter} thick, and digitised at \SI{1}{\micro\meter} in-plane resolution with a customised microscopy system -- though we downsampled them to \SI{200}{\micro\meter} to match the resolution of the MRI data (details below). We also downsampled the manual segmentations to the same resolution, and merged them into a whole brain segmentation that, after dilation, we used to mask the histological sections. The histology and associated segmentations can be interactively visualised at \url{http://atlas.brain-map.org}, and further details can be found in \citet{ding2016comprehensive}. No 3D reconstruction of the histology was performed in their study.

In addition to the histology, high-resolution MRI images of the whole brain were acquired on a 7 T Siemens scanner with a custom 30-channel receive-array coil. The specimen was scanned in a vacuum-sealed bag surrounded by Fomblin to avoid artefacts caused by  air-tissue interfaces. The images were acquired with a multiecho flash sequence (TR = 50 ms; $\alpha$ = 20$^\circ$, 40$^\circ$, 60$^\circ$, 80$^\circ$; echoes at 5.5, 12.8, 20.2, 27.6, 35.2, and 42.8 ms), at \SI{200}{\micro\meter} isotropic resolution. Once more, the details can be found in \citep{ding2016comprehensive}. In this study, we used a single volume, obtained by averaging the echoes corresponding to flip angle $\alpha$ = 20$^\circ$, which provided good contrast between gray and white matter tissue, as well as great signal-to-noise ratio. The combined image was bias field corrected with the method described in \citep{van1999automated} using the probability maps from the LONI atlas \citep{shattuck2008construction}, which  was linearly registered with NiftyReg \citep{modat2014global}. A coarse mask for the left hemisphere was manually delineated by JEI, and used to mask out tissue from the right hemisphere, which is not included in the histological analysis. Sample coronal slices of this dataset are shown in \ref{fig:exampleImages}a (histology) and \ref{fig:exampleImages}b (MRI).

\subsection{Experimental setup}
\label{sec:expSetup}

In the synthetic data, we considered three different levels of Gaussian noise ($\sigma_v=10,20,30$ mm) when generating the velocity fields, in order to model nonlinear deformations of different severity. The standard deviation of the Gaussian smoothing filter was set to 5 mm, in both the horizontal and vertical direction. The random rotations, translations and log-scalings of the similarity transform were sampled from zero-mean Gaussian distributions, with standard deviations of 2$^\circ$, 1 pixel, and 0.1, respectively. We then used NiftyReg with mutual information and our method to recover the deformations, using different spacings between control points (from 3 to 21 mm, with 3 mm steps) to evaluate different levels of model flexibility. Otherwise we used the default parameters of the package: three resolution levels, 64 bins in mutual information, and  regularisation parameters $\beta_b = 0.001$, $\beta_l=0.01$, and $\beta_j=0$. The standard deviation of the manual landmark placement was set to $\sigma_k=0.5$ mm (equivalent to 0.5 pixels). 

The rest of model parameters were set to: $\beta_1 = \beta_2 = 0.02$ (equivalent to a standard deviation of 5 mm); $a=2, b=25^2 a$ (equivalent to 4 pseudo-observations with sample standard deviation equal to 5); and $\{\bm{\Delta_s}\}$ being a  grid  covering a square with radius 10 mm, in increments of 0.5 mm. Finally, the random forest regressor consisted of 100 trees, using Gaussian derivatives and location as features. The Gaussian derivatives were of order up to three, computed at three different scales: 0, 2 and 4 mm. We grew the tree until a minimum size of 5 samples was reached at leaf nodes. We tested 5 randomly sampled features at each node. Bagging was used at both the slice and pixel levels, using 66\% of the available images, and as many pixels per image as necessary in order to have a total of 25,000 training pixels. We tested our algorithm in two different scenarios: running it on all image pairs simultaneously, or on each image pair independently (i.e., with $N=1$); the latter represents the common case that a user runs the algorithm on just a pair of images. In this case, we used 66\% of the pixels to train each tree.

In the real data, we compared mutual information based registration with our approach, using all slices simultaneously in the synthesis. In order to put the MRI in linear alignment with the histological sections, we used an iterative approach very similar to that of \citet{yang2012mri}. Starting from a stack of histological sections, we first rigidly aligned the brain MRI to the stack using mutual information. Then, we resampled the registered MR to the space of each histological section, and aligned them one by one using a similarity transform combined with mutual information. The registration of the MRI was then refined using the realigned sections, starting a new iteration. Upon convergence, we used the two competing methods to nonlinearly register the histological sections to the corresponding resampled MR images. We used the same parameters as for the experiment with the synthetic data, setting the control point spacing to the optimal values from such experiments (6 mm for the proposed approach, and 18 mm for mutual information; see Section~\ref{sec:resultsSynthetic} below); note that, for the manual landmarks, $\sigma_k=0.5$ mm was equivalent to $2.5$ pixels at the resolution of this dataset.

\subsection{Results}
\label{sec:results}

\subsubsection{Synthetic data}
\label{sec:resultsSynthetic}

\begin{figure*}
\centering
\includegraphics[width=0.82\textwidth]{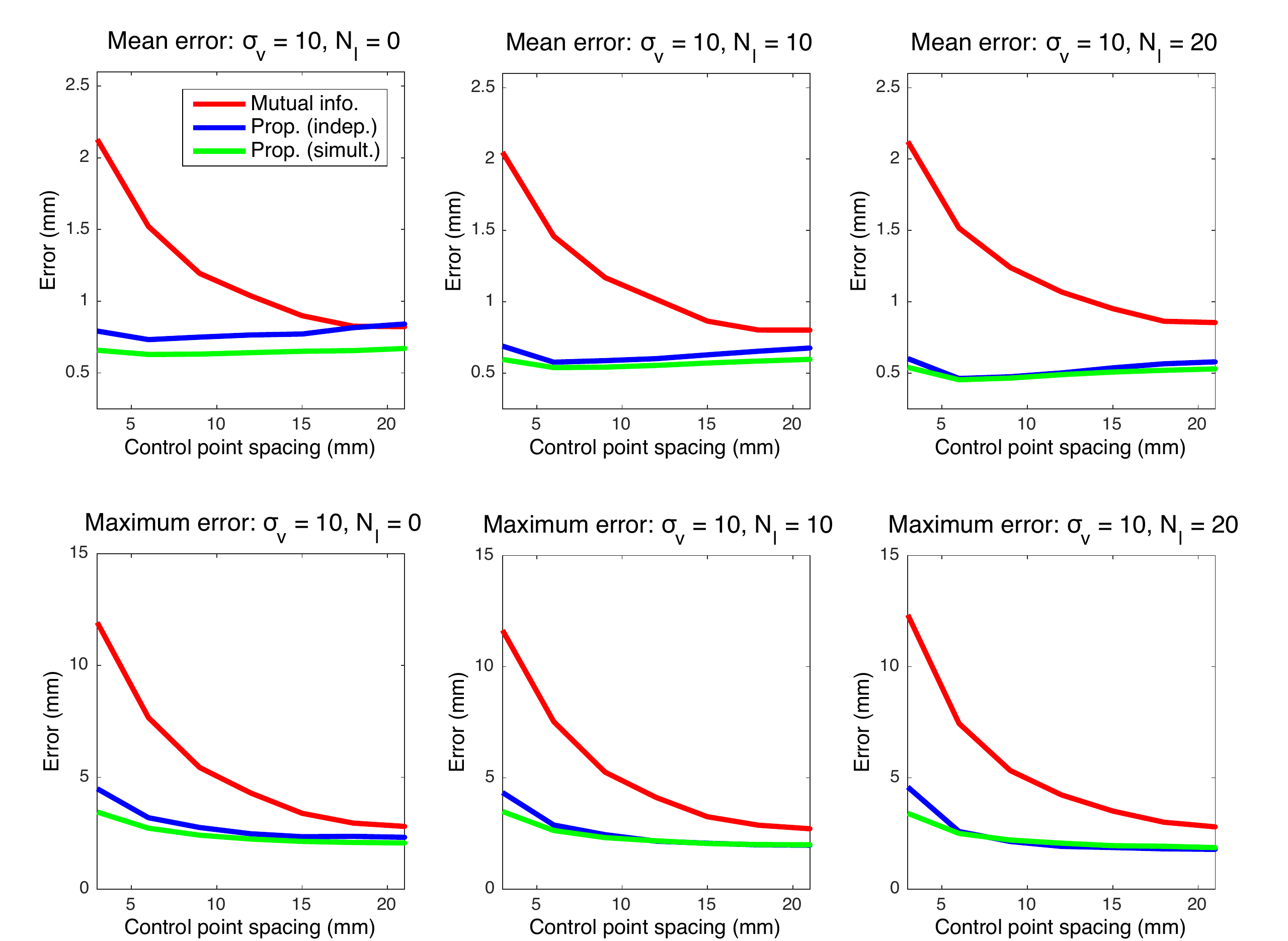}
\caption{Mean and maximum registration error in mm for deformations with $\sigma_v=10$ (mild).}
\label{fig:sigma10}
\end{figure*}

\begin{figure*}
\centering
\includegraphics[width=0.832\textwidth]{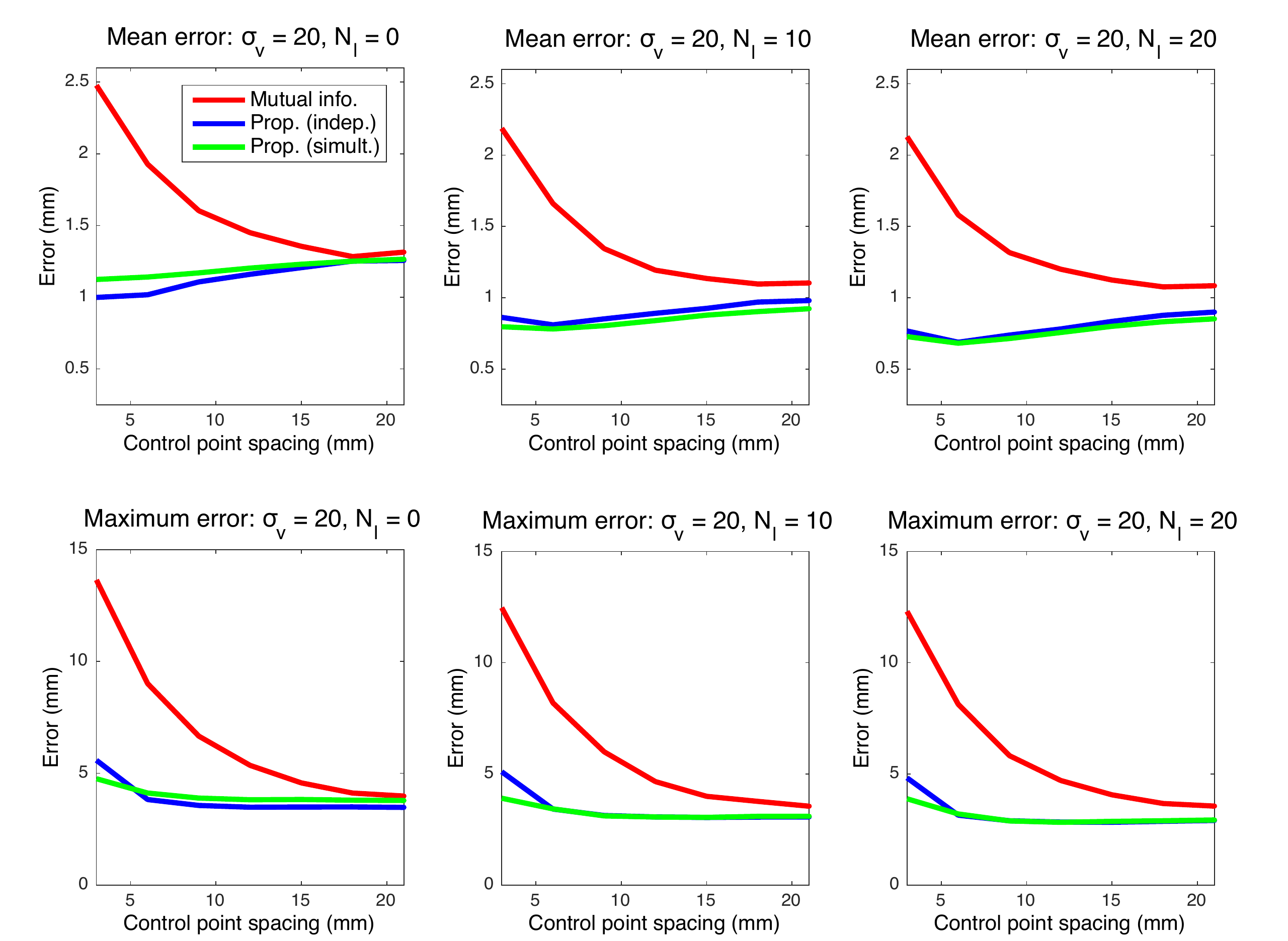}
\caption{Mean and maximum registration error in mm for deformations with $\sigma_v=20$ (medium).}
\label{fig:sigma20}
\end{figure*}

\begin{figure*}
\centering
\includegraphics[width=0.82\textwidth]{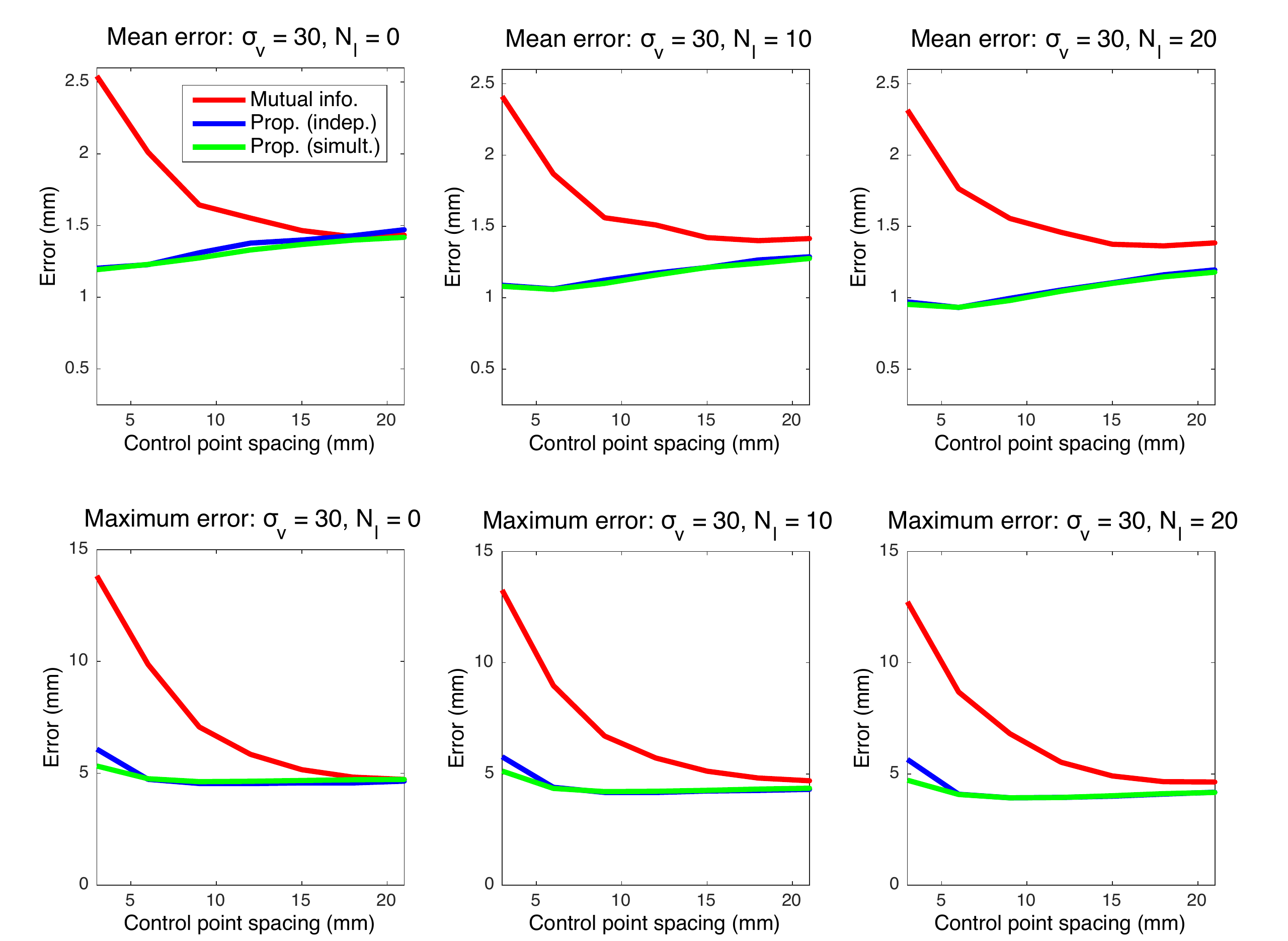}
\caption{Mean and maximum registration error in mm for deformations with $\sigma_v=10$ (strong).}
\label{fig:sigma30}
\end{figure*}

\begin{figure*}
\centering
\includegraphics[width=1.0\textwidth]{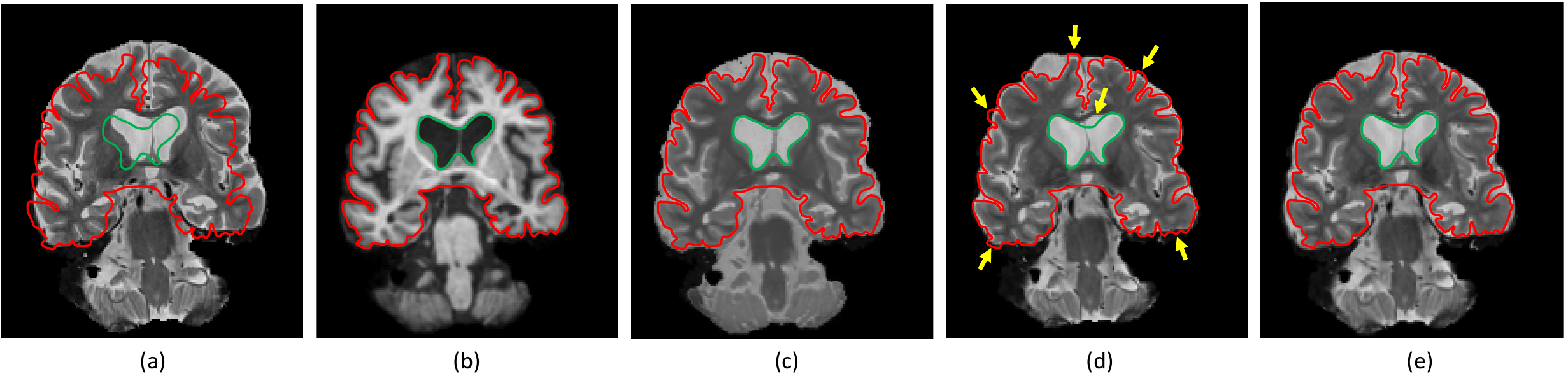}
\caption{Example from synthetic dataset: (a) Deformed T2 image, used as floating image in the registration. (b) Corresponding T1 scan, used as reference image. (c) Corresponding synthetic T2 image, after 5 iterations of our VEM algorithm. (d) Registered with mutual information. (e) Registered with our algorithm. Both in (d) and (e), the control point spacing was set to 6 mm. We have overlaid on all five images a manual outline of the gray matter surface (in red) and of the ventricles (in green), which were drawn using the T1 scan (b) as a reference. Note the poor registration produced by mutual information in the ventricles and cortical regions -- see for instance the areas pointed by the yellow arrows in (d).}
\label{fig:syntheticExamples}
\end{figure*}

Figures \ref{fig:sigma10}, \ref{fig:sigma20} and \ref{fig:sigma30} show the mean registration error as a function of the control point separation and the number of landmarks for three different levels of noise deformation: 10, 20 and 30 mm, which correspond to mild, medium and strong deformations, respectively. The mean error reflects the precision of the estimation, whereas the maximum is related to its robustness. When using mutual information, finer control point spacings in the deformation model yield transforms that are too flexible, leading to very poor results (even in presence of control points); see example in Figure~\ref{fig:syntheticExamples}. Both the mean and maximum error improve with larger spacings, flattening out at around 18-20 mm. 

The proposed method, on the other hand, provides higher precision with flexible models, thanks to the higher robustness of the intramodality metric. The two versions of the method (estimating the regressor one image pair at the time or from all images simultaneously) consistently outperform mutual information in every scenario. An important difference in the results is that the mean error hits its minimum at a much smaller control point spacing (typically 6 mm), yielding a much more accurate registration (see example in Figure~\ref{fig:syntheticExamples}). Moreover, the maximum error has already flattened at that point, in almost every tested setting.

In addition to supporting finer control points spacings, the proposed method can more effectively exploit the information provided by landmarks. In mutual information based registrations, the landmarks guide the registration, especially in the earlier iterations, since their relative cost is high. But further influence on the registration (e.g., by improving the estimation of the joint histogram) is indirect and very limited. Our proposed algorithm, on the other hand, explicitly exploits the landmark information not only in the registration, but also in the synthesis: in Equation~\ref{eq:fixedPointIts}, the landmarks create a very sharp $q$ distribution not only at pixels with landmarks, but also in the surroundings, thanks to the MRF (e.g., as in Figure~\ref{fig:exampleImages}d, Tag 1). Therefore, very similar shifted locations of these pixels are consistently selected when sampling for each tree of the forest, greatly informing the synthesis. This is reflected in the quantitative results: the gap in performance between the proposed method and mutual information widens as the number of landmarks $N_l$ increases. 

When no landmarks are used and image pairs are assessed independently, the proposed algorithm can be seen as a conventional inter-modality registration method. In that scenario, the results discussed above still hold: our method can be used at finer control point spacings, and provides average reductions of 11\%, 22\% and 15\% in the mean error, at $\sigma_v=10$, $\sigma_v=20$ and $\sigma_v=30$, respectively. We also note that, as one would expect, our method and mutual information produce almost identical results at large control point spacings.

Finally, we note a modest improvement is observed when image pairs are considered simultaneously -- rather then independently. Nevertheless, the joint estimation consistently yields higher robustness  at the finest control point spacing (3 mm), and also produces smaller errors across the different settings when the deformations are mild (Figure~\ref{fig:sigma10}). We hypothesise that, even though the simultaneous estimation has the advantage of having access to more data (which is particularly useful with more flexible models, i.e., finer spacing), the independent version can also benefit from having a regressor that is tailored to the single image pair at hand.

\subsubsection{Real data}

\label{sec:resultsReal}

\begin{figure*}
\centering
\includegraphics[width=1.0\textwidth]{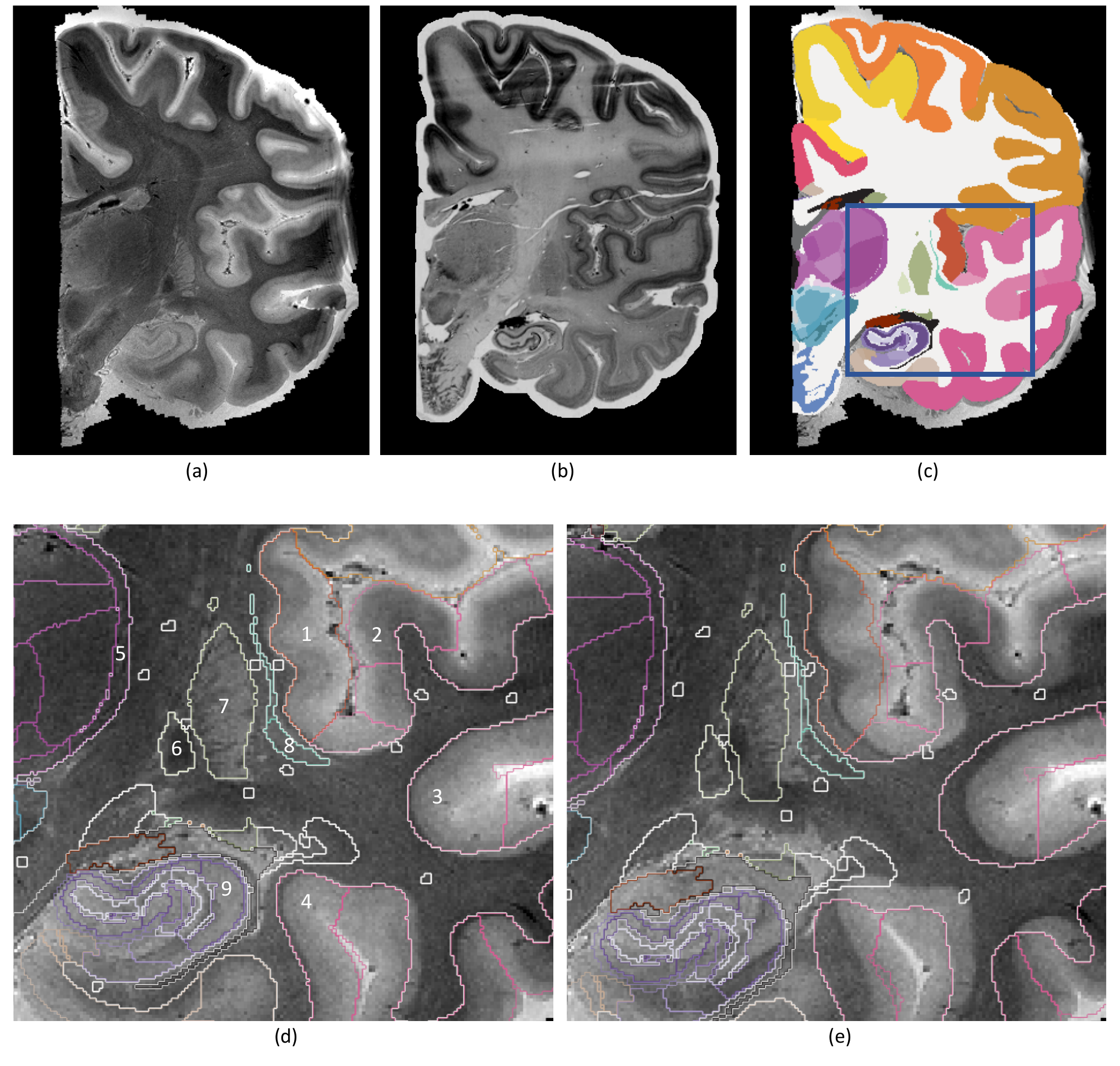}
\caption{(a) Coronal slice of the MRI scan. (b) Corresponding histological section, registered with the proposed method. (c) Corresponding manual segmentation, propagated to MR space. (d) Close-up of the region inside the blue square, showing the boundaries of the segmentation; see main text (Section~/\ref{sec:resultsReal}) for an explanation of the numerical tags. (e) Segmentation obtained when using mutual information in the registration. See \protect\url{http://atlas.brain-map.org} for the color map.}
\label{fig:AllenRes1}
\end{figure*}

Figure~\ref{fig:AllenRes1} shows a representative coronal section of the data, which covers multiple cortical and subcortical structures of interest (e.g., hippocampus, thalamus, putamen and pallidum). Comparing the segmentations propagated from the histology to the MRI with the proposed method (Figure~\ref{fig:AllenRes1}d) and mutual information (Figure~\ref{fig:AllenRes1}e), it is apparent that our algorithm produces a much more accurate registration. The contours of the white matter surface are rather inaccurate when using mutual information; see for instance the insular (Tag 1 in the figure), auditory (Tag 2), or  polysensoral temporal cortices (Tag 3); or area 36 (Tag 4).  Using the proposed method, the registered contours follow the underlying MRI intensities much more accurately. The same applies to subcortical structures. In the thalamus (light purple), it can be seen that the segmentation of the reticular nucleus (Tag 5) is too medial when using mutual information. The same applies to the pallidum (Tag 6), putamen (Tag 7) and claustrum (Tag 8). The hippocampus (dark purple; Tag 9) is  too inferior to the actual anatomy in the MRI. Once more, the proposed algorithm produces, qualitatively speaking, much improved boundaries. 

\begin{figure*}
\centering
\includegraphics[width=0.84\textwidth]{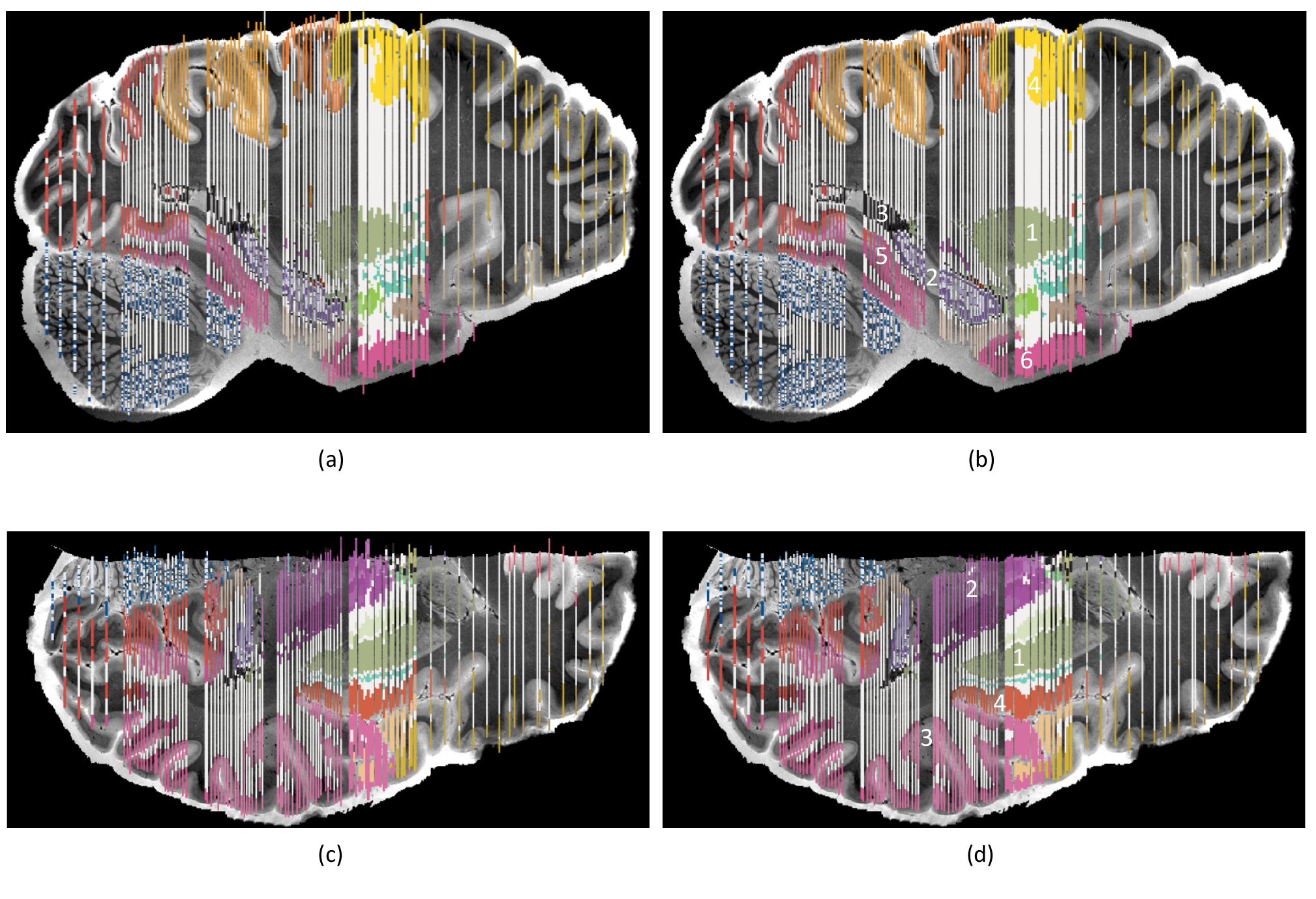}
\caption{(a) Sagittal slice of the MRI scan, with registered segmentation superimposed. The deformation fields used to propagate the manual segmentations from histology to MRI space were computed with mutual information. (b) Same as (a), but using our technique to register the data. (c) Axial slice, reconstruction with mutual information. (d) Same slice, reconstructed with our proposed method. See \protect\url{http://atlas.brain-map.org} for the color map.}
\label{fig:AllenRes2}
\end{figure*}

To better assess the quality of the reconstruction as a whole (rather than on a single slice), Figure~\ref{fig:AllenRes2} shows the propagated segmentations in the orthogonal views: sagittal (Figures \ref{fig:AllenRes2}a, \ref{fig:AllenRes2}b) and axial (Figures \ref{fig:AllenRes2}c, \ref{fig:AllenRes2}d). The proposed method produces reconstructed segmentations that are smoother and that better follow the anatomy in the MRI scan. In sagittal view, this can be easily observed in subcortical regions such as the putamen (Tag 1 in Figure~\ref{fig:AllenRes2}b), the hippocampus (Tag 2) or the lateral ventricle (Tag 3); and also in cortical regions such as the premotor (Tag 4), parahippocampal (Tag 5) or fusiform temporal (Tag 6) cortices. The improvement is also apparent from how much less frequently the segmentation leaks outside the brain when using our algorithm. Similar conclusions can be derived from the axial view; see for instance the putamen (Tag 1 in Figure~\ref{fig:AllenRes2}d), thalamus (purple region, Tag 2), polysensory temporal cortex (Tag 3) or insular cortex (Tag 4).


\section{Discussion and conclusion}
\label{sec:discussion}
 
In this article, we presented a novel method to simultaneously estimate the registration and synthesis between a pair of corresponding images from different modalities. The results on both synthetic (quantitative) and real data (qualitative) show that the proposed algorithm is superior to standard inter-modality registration based on mutual information, albeit slower due to the need to iterate between registration and synthesis -- especially the former, since it requires nested iteration of Equation~\ref{eq:fixedPointIts}. Our Matlab implementation runs in 2-3 minutes for images of size $256^2$ pixels, but parallelised implementation in C++  or on the GPU should greatly reduce the running time. 

The quantitative experiments demonstrated that our algorithm supports much more flexible deformation models than mutual informations (i.e., smaller control point spacing) without compromising robustness, attributed to the more stable intra-modality metric (which we have made publicly available in NiftyReg). Moreover, the experiments on synthetic data also showed that our algorithm can more effectively take advantage of the information encoded in manually placed pairs of landmarks, since this can be exploited in both the registration and synthesis, which inform each other in the model fitting. The more landmarks we used, the larger the gap between our method and mutual information was -- however, we should note that, in the limit, the performance of the two methods would be the same, since the registration error would go to zero in both cases. 

We must note that, in the experiments with synthetic data, the relative contributions of the data terms in Equations \ref{eq:costRegNiftyReg}) and \ref{eq:costRegMIinNiftyReg} are slightly different, since computing the value of $\alpha$ that makes these contributions exactly equal is very difficult. However, the minor differences that our heuristic choice of $\alpha$ might introduce do not undermine the results of the experiments, since the approximate effect of modifying $\alpha$ is mildly shifting the curves in Figures~\ref{fig:sigma10}-\ref{fig:sigma30} to the left or right -- which does not change the conclusions. 

Our method also outperformed mutual information when applied to the data from the Allen Institute, which is more challenging due to the more complex relationships between the two contrast mechanisms, and the presence of artefacts such as cracks and tears. Qualitatively speaking, the superiority of our approach is clearly apparent from Figure~\ref{fig:AllenRes2}, in which it produces a much smoother segmentation in the orthogonal planes. We note that we did not introduce any smoothness constraints in the reconstruction, e.g., by forcing the registered histological sections to be similar to their neighbours, through an explicit term in the cost function of the registration. Such a strategy would produce smoother reconstructions, but these would not necessarily be more accurate -- particularly if one considers that the 2D deformations fields of the different sections are independent \emph{a priori}, which makes the histological sections conditionally independent \emph{a posteriori}, given the MRI data and the image intensity transform. Moreover, explicitly enforcing such smoothness in the registration would preclude qualitative evaluation through visual inspection of the segmentation in the orthogonal orientations.

The proposed algorithm is hybrid in the sense that, despite being formulated in a generative framework, it replaces  the likelihood term of the synthesis by a discriminative element. We emphasise that such a change still yields a valid objective function (Equation~\ref{eq:objectiveFun}) that we can approximately optimise with VEM -- which maximises Equations \ref{eq:targetE} and \ref{eq:targetM} instead. The VEM algorithm alternately optimises for $q$ and $\theta$ in a coordinate descent scheme, and is in principle guaranteed to converge. In our method, we lose this property due to the approximate optimisation of the random forest parameters, since injecting randomness is one of the key elements of the success of random decision trees. However, in practice, our algorithm typically converges in 5-6 iterations, in terms of changes in the predicted synthetic image (i.e., in $\mu_{n\bm{x}}$ and $\sigma^2_{n\bm{x}}$).

Our approach can also be used in an online manner, i.e., if data become progressively available at testing. For example, the random forest could be optimised on an (ideally) large set of images, considering them simultaneously in the framework. Then, when a new pair of images arrives, one can assume that the forest parameters are fixed and equal to $\hat{\theta}$, and proceed directly to the estimation of the synthetic image $\mu_{1\bm{x}}, \sigma^2_{1\bm{x}}$ and deformation field $\hat{\bm{U}}_1$. An alternative would  be to fine tune $\theta$ to the new input, considering it in isolation or jointly with the other scans. But even if no other previous data are available (i.e., $N=1$), the registration uncertainty encoded in $q$ prevents the regression from overfitting, and enables our method to still outperform mutual information. This is in contrast with supervised synthesis algorithms, which cannot operate without training data. 

Future work will follow four main directions. First, integrating deep learning techniques into the framework, which could be particularly useful when large amounts of image pairs are available, e.g., in a large histology reconstruction project. The main challenges to tackle are overfitting and avoiding to make the algorithm impractically slow. A possible solution to this problem would be to use a pretrained network, and only update the connections in the last layer during the analysis of the image pair at hand (e.g., as in \citealt{wang2017interactive}). A second direction of future work is the extension of the algorithm to 3D. Albeit mathematically straightforward (no changes are required in the framework),  such extension poses problems from the practical perspective, e.g., the memory requirements for storing $q$ grow very quickly. A third avenue of future work is the application to other target modalities, such as optical coherence tomography (OCT). Finally, we will also explore the possibility of synthesizing histology from MRI. This a more challenging task that might require multiple input MRI contrasts, depending on the target stain to synthesise. However, synthetic histology would not only provide an estimate of the microanatomy of tissue imaged with MRI, but would also enable the symmetrisation of the framework presented in this article; by computing two syntheses, the robustness of the algorithm would be be expected to increase.

The algorithm presented in this paper represents a significant step towards solving the problem of aligning histological images and MRI, by exploiting the connection between registration and synthesis within a novel probabilistic framework. We will use this method to produce increasingly precise histological reconstructions of tissue, which in turn will enable us to build probabilistic atlases of the human brain at a superior level of detail.


\section*{Acknowledgements}
This research was supported by the European Research Council (ERC) Starting Grant No. 677697 (project BUNGEE-TOOLS), and from the Wellcome Trust and EPSRC (203145Z/16/Z, WT101957, NS/A000027/1, NS/A000050/1). 
Support for this research was also provided in part by the National Institute for Biomedical
Imaging and Bioengineering (P41EB015896, 1R01EB023281,
R01EB006758, R21EB018907,
R01EB019956), the National Institute on Aging (5R01AG008122,
R01AG016495), the National Institute of Diabetes and Digestive and
Kidney Diseases (1-R21-DK-108277-01), the National Institute for
Neurological Disorders and Stroke (R01NS0525851, R21NS072652,
R01NS070963, R01NS083534, 5U01NS086625), and was made possible by the
resources provided by Shared Instrumentation Grants 1S10RR023401,
1S10RR019307, and 1S10RR023043. Additional support was provided by the NIH
Blueprint for Neuroscience Research (5U01-MH093765), part of the
multi-institutional Human Connectome Project. In addition, BF has a
financial interest in CorticoMetrics, a company whose medical pursuits
focus on brain imaging and measurement technologies. BF's interests were
reviewed and are managed by Massachusetts General Hospital and Partners
HealthCare in accordance with their conflict of interest policies.

The collection and sharing of the MRI data used in the group study based on ADNI was funded by the Alzheimer's Disease Neuroimaging Initiative (National Institutes of Health Grant U01 AG024904) and DOD ADNI (Department of Defense award number W81XWH-12-2-0012). ADNI is funded by the
National Institute on Aging, the National Institute of Biomedical Imaging and Bioengineering, and through generous contributions from the following: Alzheimer's Association; Alzheimer's Drug Discovery Foundation; BioClinica, Inc.; Biogen Idec Inc.; Bristol-Myers Squibb Company; Eisai Inc.; Elan Pharmaceuticals, Inc.; Eli Lilly and Company; F. Hoffmann-La Roche Ltd and its affiliated company Genentech, Inc.; GE Healthcare; Innogenetics, N.V.; IXICO Ltd.; Janssen Alzheimer Immunotherapy Research \& Development, LLC.; Johnson \& Johnson Pharmaceutical Research \&
Development LLC.; Medpace, Inc.; Merck \& Co., Inc.; Meso Scale Diagnostics, LLC.; NeuroRx Research; Novartis Pharmaceuticals Corporation; Pfizer Inc.; Piramal Imaging; Servier; Synarc Inc.; and Takeda Pharmaceutical Company. The Canadian Institutes of Health Research is providing funds to support ADNI clinical sites in Canada. Private sector contributions are facilitated by the Foundation for the National Institutes of Health (www.fnih.org). The grantee organisation is the Northern California Institute for Research and Education, and the study is coordinated by the Alzheimer's Disease Cooperative Study at the University of California, San Diego. ADNI data are disseminated by the Laboratory for Neuro Imaging at the University of Southern California.


\bibliographystyle{model2-names}
\bibliography{bibliography.bib}

\begin{thebibliography}{67}
\expandafter\ifx\csname natexlab\endcsname\relax\def\natexlab#1{#1}\fi
\expandafter\ifx\csname url\endcsname\relax
  \def\url#1{\texttt{#1}}\fi
\expandafter\ifx\csname urlprefix\endcsname\relax\def\urlprefix{URL }\fi
\providecommand{\eprint}[2][]{\url{#2}}
\providecommand{\bibinfo}[2]{#2}
\ifx\xfnm\relax \def\xfnm[#1]{\unskip,\space#1}\fi
\bibitem[{Adler et~al.(2016)Adler, Ittyerah, Pluta, Pickup, Liu, Wolk and
  Yushkevich}]{adler2016probabilistic}
\bibinfo{author}{Adler, D.H.}, \bibinfo{author}{Ittyerah, R.},
  \bibinfo{author}{Pluta, J.}, \bibinfo{author}{Pickup, S.},
  \bibinfo{author}{Liu, W.}, \bibinfo{author}{Wolk, D.A.},
  \bibinfo{author}{Yushkevich, P.A.}, \bibinfo{year}{2016}.
\newblock \bibinfo{title}{Probabilistic atlas of the human hippocampus
  combining ex vivo mri and histology}, in: \bibinfo{booktitle}{International
  Conference on Medical Image Computing and Computer-Assisted Intervention},
  \bibinfo{organization}{Springer}. pp. \bibinfo{pages}{63--71}.
\bibitem[{Adler et~al.(2014)Adler, Pluta, Kadivar, Craige, Gee, Avants and
  Yushkevich}]{adler2014histology}
\bibinfo{author}{Adler, D.H.}, \bibinfo{author}{Pluta, J.},
  \bibinfo{author}{Kadivar, S.}, \bibinfo{author}{Craige, C.},
  \bibinfo{author}{Gee, J.C.}, \bibinfo{author}{Avants, B.B.},
  \bibinfo{author}{Yushkevich, P.A.}, \bibinfo{year}{2014}.
\newblock \bibinfo{title}{Histology-derived volumetric annotation of the human
  hippocampal subfields in postmortem {MRI}}.
\newblock \bibinfo{journal}{Neuroimage} \bibinfo{volume}{84},
  \bibinfo{pages}{505--523}.
\bibitem[{Ahuja et~al.(1993)Ahuja, Magnanti and Orlin}]{ahuja1993network}
\bibinfo{author}{Ahuja, R.K.}, \bibinfo{author}{Magnanti, T.L.},
  \bibinfo{author}{Orlin, J.B.}, \bibinfo{year}{1993}.
\newblock \bibinfo{title}{Network flows: theory, algorithms, and applications}
  .
\bibitem[{Amunts et~al.(2013)Amunts, Lepage, Borgeat, Mohlberg, Dickscheid,
  Rousseau, Bludau, Bazin, Lewis, Oros-Peusquens, Shah, Lippert, Zilles and
  Evans}]{amunts2013bigbrain}
\bibinfo{author}{Amunts, K.}, \bibinfo{author}{Lepage, C.},
  \bibinfo{author}{Borgeat, L.}, \bibinfo{author}{Mohlberg, H.},
  \bibinfo{author}{Dickscheid, T.}, \bibinfo{author}{Rousseau, M.{\'E}.},
  \bibinfo{author}{Bludau, S.}, \bibinfo{author}{Bazin, P.L.},
  \bibinfo{author}{Lewis, L.B.}, \bibinfo{author}{Oros-Peusquens, A.M.},
  \bibinfo{author}{Shah, N.J.}, \bibinfo{author}{Lippert, T.},
  \bibinfo{author}{Zilles, K.}, \bibinfo{author}{Evans, A.C.},
  \bibinfo{year}{2013}.
\newblock \bibinfo{title}{Bigbrain: an ultrahigh-resolution 3d human brain
  model}.
\newblock \bibinfo{journal}{Science} \bibinfo{volume}{340},
  \bibinfo{pages}{1472--1475}.
\bibitem[{Arganda-Carreras et~al.(2010)Arganda-Carreras,
  Fern{\'a}ndez-Gonz{\'a}lez, Mu{\~n}oz-Barrutia and
  Ortiz-De-Solorzano}]{arganda20103d}
\bibinfo{author}{Arganda-Carreras, I.},
  \bibinfo{author}{Fern{\'a}ndez-Gonz{\'a}lez, R.},
  \bibinfo{author}{Mu{\~n}oz-Barrutia, A.},
  \bibinfo{author}{Ortiz-De-Solorzano, C.}, \bibinfo{year}{2010}.
\newblock \bibinfo{title}{3d reconstruction of histological sections:
  application to mammary gland tissue}.
\newblock \bibinfo{journal}{Microscopy research and technique}
  \bibinfo{volume}{73}, \bibinfo{pages}{1019--1029}.
\bibitem[{Arsigny et~al.(2006)Arsigny, Commowick, Pennec and
  Ayache}]{arsigny2006log}
\bibinfo{author}{Arsigny, V.}, \bibinfo{author}{Commowick, O.},
  \bibinfo{author}{Pennec, X.}, \bibinfo{author}{Ayache, N.},
  \bibinfo{year}{2006}.
\newblock \bibinfo{title}{A log-euclidean framework for statistics on
  diffeomorphisms}.
\newblock \bibinfo{journal}{Medical Image Computing and Computer-Assisted
  Intervention--MICCAI 2006} , \bibinfo{pages}{924--931}.
\bibitem[{Augustinack et~al.(2005)Augustinack, van~der Kouwe, Blackwell, Salat,
  Wiggins, Frosch, Wiggins, Potthast, Wald and
  Fischl}]{augustinack2005detection}
\bibinfo{author}{Augustinack, J.C.}, \bibinfo{author}{van~der Kouwe, A.J.},
  \bibinfo{author}{Blackwell, M.L.}, \bibinfo{author}{Salat, D.H.},
  \bibinfo{author}{Wiggins, C.J.}, \bibinfo{author}{Frosch, M.P.},
  \bibinfo{author}{Wiggins, G.C.}, \bibinfo{author}{Potthast, A.},
  \bibinfo{author}{Wald, L.L.}, \bibinfo{author}{Fischl, B.R.},
  \bibinfo{year}{2005}.
\newblock \bibinfo{title}{Detection of entorhinal layer ii using tesla magnetic
  resonance imaging}.
\newblock \bibinfo{journal}{Annals of neurology} \bibinfo{volume}{57},
  \bibinfo{pages}{489--494}.
\bibitem[{Boykov et~al.(2001)Boykov, Veksler and Zabih}]{boykov2001fast}
\bibinfo{author}{Boykov, Y.}, \bibinfo{author}{Veksler, O.},
  \bibinfo{author}{Zabih, R.}, \bibinfo{year}{2001}.
\newblock \bibinfo{title}{Fast approximate energy minimization via graph cuts}.
\newblock \bibinfo{journal}{IEEE Transactions on pattern analysis and machine
  intelligence} \bibinfo{volume}{23}, \bibinfo{pages}{1222--1239}.
\bibitem[{Breiman(1996)}]{breiman1996bagging}
\bibinfo{author}{Breiman, L.}, \bibinfo{year}{1996}.
\newblock \bibinfo{title}{Bagging predictors}.
\newblock \bibinfo{journal}{Machine learning} \bibinfo{volume}{24},
  \bibinfo{pages}{123--140}.
\bibitem[{Breiman(2001)}]{breiman2001random}
\bibinfo{author}{Breiman, L.}, \bibinfo{year}{2001}.
\newblock \bibinfo{title}{Random forests}.
\newblock \bibinfo{journal}{Machine learning} \bibinfo{volume}{45},
  \bibinfo{pages}{5--32}.
\bibitem[{Burgos et~al.(2014)Burgos, Cardoso, Thielemans, Modat, Pedemonte,
  Dickson, Barnes, Ahmed, Mahoney, Schott, Duncan, Atkinson, Arridge, Hutton
  and Ourselin}]{burgos2014attenuation}
\bibinfo{author}{Burgos, N.}, \bibinfo{author}{Cardoso, M.J.},
  \bibinfo{author}{Thielemans, K.}, \bibinfo{author}{Modat, M.},
  \bibinfo{author}{Pedemonte, S.}, \bibinfo{author}{Dickson, J.},
  \bibinfo{author}{Barnes, A.}, \bibinfo{author}{Ahmed, R.},
  \bibinfo{author}{Mahoney, C.J.}, \bibinfo{author}{Schott, J.M.},
  \bibinfo{author}{Duncan, J.S.}, \bibinfo{author}{Atkinson, D.},
  \bibinfo{author}{Arridge, S.R.}, \bibinfo{author}{Hutton, B.F.},
  \bibinfo{author}{Ourselin, S.}, \bibinfo{year}{2014}.
\newblock \bibinfo{title}{Attenuation correction synthesis for hybrid
  {PET}-{MR} scanners: application to brain studies}.
\newblock \bibinfo{journal}{IEEE transactions on medical imaging}
  \bibinfo{volume}{33}, \bibinfo{pages}{2332--2341}.
\bibitem[{Cachier et~al.(2003)Cachier, Bardinet, Dormont, Pennec and
  Ayache}]{cachier2003iconic}
\bibinfo{author}{Cachier, P.}, \bibinfo{author}{Bardinet, E.},
  \bibinfo{author}{Dormont, D.}, \bibinfo{author}{Pennec, X.},
  \bibinfo{author}{Ayache, N.}, \bibinfo{year}{2003}.
\newblock \bibinfo{title}{Iconic feature based nonrigid registration: the
  {PASHA} algorithm}.
\newblock \bibinfo{journal}{Computer vision and image understanding}
  \bibinfo{volume}{89}, \bibinfo{pages}{272--298}.
\bibitem[{Chakravarty et~al.(2006)Chakravarty, Bertrand, Hodge, Sadikot and
  Collins}]{chakravarty2006creation}
\bibinfo{author}{Chakravarty, M.M.}, \bibinfo{author}{Bertrand, G.},
  \bibinfo{author}{Hodge, C.P.}, \bibinfo{author}{Sadikot, A.F.},
  \bibinfo{author}{Collins, D.L.}, \bibinfo{year}{2006}.
\newblock \bibinfo{title}{The creation of a brain atlas for image guided
  neurosurgery using serial histological data}.
\newblock \bibinfo{journal}{Neuroimage} \bibinfo{volume}{30},
  \bibinfo{pages}{359--376}.
\bibitem[{Chartsias et~al.(2017)Chartsias, Joyce, Dharmakumar and
  Tsaftaris}]{chartsias2017adversarial}
\bibinfo{author}{Chartsias, A.}, \bibinfo{author}{Joyce, T.},
  \bibinfo{author}{Dharmakumar, R.}, \bibinfo{author}{Tsaftaris, S.A.},
  \bibinfo{year}{2017}.
\newblock \bibinfo{title}{Adversarial image synthesis for unpaired multi-modal
  cardiac data}, in: \bibinfo{booktitle}{International Workshop on Simulation
  and Synthesis in Medical Imaging}, \bibinfo{organization}{Springer}. pp.
  \bibinfo{pages}{3--13}.
\bibitem[{Cifor et~al.(2011)Cifor, Bai and Pitiot}]{cifor2011smoothness}
\bibinfo{author}{Cifor, A.}, \bibinfo{author}{Bai, L.},
  \bibinfo{author}{Pitiot, A.}, \bibinfo{year}{2011}.
\newblock \bibinfo{title}{Smoothness-guided 3-d reconstruction of 2-d
  histological images}.
\newblock \bibinfo{journal}{NeuroImage} \bibinfo{volume}{56},
  \bibinfo{pages}{197--211}.
\bibitem[{Collins et~al.(1994)Collins, Neelin, Peters and
  Evans}]{collins1994automatic}
\bibinfo{author}{Collins, D.L.}, \bibinfo{author}{Neelin, P.},
  \bibinfo{author}{Peters, T.M.}, \bibinfo{author}{Evans, A.C.},
  \bibinfo{year}{1994}.
\newblock \bibinfo{title}{Automatic {3D} intersubject registration of {MR}
  volumetric data in standardized talairach space.}
\newblock \bibinfo{journal}{Journal of computer assisted tomography}
  \bibinfo{volume}{18}, \bibinfo{pages}{192--205}.
\bibitem[{Criminisi et~al.(2011)Criminisi, Shotton and
  Konukoglu}]{criminisi2011decision}
\bibinfo{author}{Criminisi, A.}, \bibinfo{author}{Shotton, J.},
  \bibinfo{author}{Konukoglu, E.}, \bibinfo{year}{2011}.
\newblock \bibinfo{title}{Decision forests for classification, regression,
  density estimation, manifold learning and semi-supervised learning}.
\newblock \bibinfo{journal}{Microsoft Research Cambridge, Tech. Rep.
  MSRTR-2011-114} \bibinfo{volume}{5}, \bibinfo{pages}{12}.
\bibitem[{Dauguet et~al.(2007)Dauguet, Delzescaux, Cond{\'e}, Mangin, Ayache,
  Hantraye and Frouin}]{dauguet2007three}
\bibinfo{author}{Dauguet, J.}, \bibinfo{author}{Delzescaux, T.},
  \bibinfo{author}{Cond{\'e}, F.}, \bibinfo{author}{Mangin, J.F.},
  \bibinfo{author}{Ayache, N.}, \bibinfo{author}{Hantraye, P.},
  \bibinfo{author}{Frouin, V.}, \bibinfo{year}{2007}.
\newblock \bibinfo{title}{Three-dimensional reconstruction of stained
  histological slices and 3d non-linear registration with in-vivo mri for whole
  baboon brain}.
\newblock \bibinfo{journal}{Journal of neuroscience methods}
  \bibinfo{volume}{164}, \bibinfo{pages}{191--204}.
\bibitem[{Ding et~al.(2016)Ding, Royall, Sunkin, Ng, Facer, Lesnar,
  Guillozet-Bongaarts, McMurray, Szafer, Dolbeare, Stevens, Tirrell, Benner,
  Caldejon, Dalley, Dee, Lau, Nyhus, Reding, Riley, Sandman, Shen, van~der
  Kouwe, Varjabedian, Write, Zollei, Dang, Knowles, Koch, Phillips, Sestan,
  Wohnoutka, Zielke, Hohmann, Jones, Bernard, Hawrylycz, Hof, Fischl and
  Lein}]{ding2016comprehensive}
\bibinfo{author}{Ding, S.L.}, \bibinfo{author}{Royall, J.J.},
  \bibinfo{author}{Sunkin, S.M.}, \bibinfo{author}{Ng, L.},
  \bibinfo{author}{Facer, B.A.}, \bibinfo{author}{Lesnar, P.},
  \bibinfo{author}{Guillozet-Bongaarts, A.}, \bibinfo{author}{McMurray, B.},
  \bibinfo{author}{Szafer, A.}, \bibinfo{author}{Dolbeare, T.A.},
  \bibinfo{author}{Stevens, A.}, \bibinfo{author}{Tirrell, L.},
  \bibinfo{author}{Benner, T.}, \bibinfo{author}{Caldejon, S.},
  \bibinfo{author}{Dalley, R.A.}, \bibinfo{author}{Dee, N.},
  \bibinfo{author}{Lau, C.}, \bibinfo{author}{Nyhus, J.},
  \bibinfo{author}{Reding, M.}, \bibinfo{author}{Riley, Z.L.},
  \bibinfo{author}{Sandman, D.}, \bibinfo{author}{Shen, E.},
  \bibinfo{author}{van~der Kouwe, A.}, \bibinfo{author}{Varjabedian, A.},
  \bibinfo{author}{Write, M.}, \bibinfo{author}{Zollei, L.},
  \bibinfo{author}{Dang, C.}, \bibinfo{author}{Knowles, J.A.},
  \bibinfo{author}{Koch, C.}, \bibinfo{author}{Phillips, J.W.},
  \bibinfo{author}{Sestan, N.}, \bibinfo{author}{Wohnoutka, P.},
  \bibinfo{author}{Zielke, H.R.}, \bibinfo{author}{Hohmann, J.G.},
  \bibinfo{author}{Jones, A.R.}, \bibinfo{author}{Bernard, A.},
  \bibinfo{author}{Hawrylycz, M.J.}, \bibinfo{author}{Hof, P.R.},
  \bibinfo{author}{Fischl, B.}, \bibinfo{author}{Lein, E.S.},
  \bibinfo{year}{2016}.
\newblock \bibinfo{title}{Comprehensive cellular-resolution atlas of the adult
  human brain}.
\newblock \bibinfo{journal}{Journal of Comparative Neurology}
  \bibinfo{volume}{524}, \bibinfo{pages}{3127--3481}.
\bibitem[{Ebner et~al.(2017)Ebner, Chung, Prados, Cardoso, Chard, Vercauteren
  and Ourselin}]{ebner2017volumetric}
\bibinfo{author}{Ebner, M.}, \bibinfo{author}{Chung, K.K.},
  \bibinfo{author}{Prados, F.}, \bibinfo{author}{Cardoso, M.J.},
  \bibinfo{author}{Chard, D.T.}, \bibinfo{author}{Vercauteren, T.},
  \bibinfo{author}{Ourselin, S.}, \bibinfo{year}{2017}.
\newblock \bibinfo{title}{Volumetric reconstruction from printed films:
  Enabling 30 year longitudinal analysis in mr neuroimaging}.
\newblock \bibinfo{journal}{NeuroImage} \bibinfo{volume}{165},
  \bibinfo{pages}{238--250}.
\bibitem[{Evans et~al.(1993)Evans, Collins, Mills, Brown, Kelly and
  Peters}]{evans19933d}
\bibinfo{author}{Evans, A.C.}, \bibinfo{author}{Collins, D.L.},
  \bibinfo{author}{Mills, S.}, \bibinfo{author}{Brown, E.},
  \bibinfo{author}{Kelly, R.}, \bibinfo{author}{Peters, T.M.},
  \bibinfo{year}{1993}.
\newblock \bibinfo{title}{{3D} statistical neuroanatomical models from 305
  {MRI} volumes}, in: \bibinfo{booktitle}{Nuclear Science Symposium and Medical
  Imaging Conference, 1993., 1993 IEEE Conference Record.},
  \bibinfo{organization}{IEEE}. pp. \bibinfo{pages}{1813--1817}.
\bibitem[{Fischl(2012)}]{fischl2012freesurfer}
\bibinfo{author}{Fischl, B.}, \bibinfo{year}{2012}.
\newblock \bibinfo{title}{Freesurfer}.
\newblock \bibinfo{journal}{Neuroimage} \bibinfo{volume}{62},
  \bibinfo{pages}{774--781}.
\bibitem[{Fischl et~al.(2002)Fischl, Salat, Busa, Albert, Dieterich,
  Haselgrove, Van Der~Kouwe, Killiany, Kennedy, Klaveness, Montillo, Makris,
  Rosen and Dale}]{fischl2002whole}
\bibinfo{author}{Fischl, B.}, \bibinfo{author}{Salat, D.H.},
  \bibinfo{author}{Busa, E.}, \bibinfo{author}{Albert, M.},
  \bibinfo{author}{Dieterich, M.}, \bibinfo{author}{Haselgrove, C.},
  \bibinfo{author}{Van Der~Kouwe, A.}, \bibinfo{author}{Killiany, R.},
  \bibinfo{author}{Kennedy, D.}, \bibinfo{author}{Klaveness, S.},
  \bibinfo{author}{Montillo, A.}, \bibinfo{author}{Makris, N.},
  \bibinfo{author}{Rosen, B.}, \bibinfo{author}{Dale, A.M.},
  \bibinfo{year}{2002}.
\newblock \bibinfo{title}{Whole brain segmentation: automated labeling of
  neuroanatomical structures in the human brain}.
\newblock \bibinfo{journal}{Neuron} \bibinfo{volume}{33},
  \bibinfo{pages}{341--355}.
\bibitem[{Harris and Stephens(1988)}]{harris1988combined}
\bibinfo{author}{Harris, C.}, \bibinfo{author}{Stephens, M.},
  \bibinfo{year}{1988}.
\newblock \bibinfo{title}{A combined corner and edge detector.}, in:
  \bibinfo{booktitle}{Alvey vision conference},
  \bibinfo{organization}{Manchester, UK}. pp. \bibinfo{pages}{10--5244}.
\bibitem[{Hertzmann et~al.(2001)Hertzmann, Jacobs, Oliver, Curless and
  Salesin}]{hertzmann2001image}
\bibinfo{author}{Hertzmann, A.}, \bibinfo{author}{Jacobs, C.E.},
  \bibinfo{author}{Oliver, N.}, \bibinfo{author}{Curless, B.},
  \bibinfo{author}{Salesin, D.H.}, \bibinfo{year}{2001}.
\newblock \bibinfo{title}{Image analogies}, in: \bibinfo{booktitle}{Proceedings
  of the 28th annual conference on Computer graphics and interactive
  techniques}, \bibinfo{organization}{ACM}. pp. \bibinfo{pages}{327--340}.
\bibitem[{Hibbard and Hawkins(1988)}]{hibbard1988objective}
\bibinfo{author}{Hibbard, L.S.}, \bibinfo{author}{Hawkins, R.A.},
  \bibinfo{year}{1988}.
\newblock \bibinfo{title}{Objective image alignment for three-dimensional
  reconstruction of digital autoradiograms}.
\newblock \bibinfo{journal}{Journal of neuroscience methods}
  \bibinfo{volume}{26}, \bibinfo{pages}{55--74}.
\bibitem[{Holmes et~al.(1998)Holmes, Hoge, Collins, Woods, Toga and
  Evans}]{holmes1998enhancement}
\bibinfo{author}{Holmes, C.J.}, \bibinfo{author}{Hoge, R.},
  \bibinfo{author}{Collins, L.}, \bibinfo{author}{Woods, R.},
  \bibinfo{author}{Toga, A.W.}, \bibinfo{author}{Evans, A.C.},
  \bibinfo{year}{1998}.
\newblock \bibinfo{title}{Enhancement of {MR} images using registration for
  signal averaging}.
\newblock \bibinfo{journal}{Journal of computer assisted tomography}
  \bibinfo{volume}{22}, \bibinfo{pages}{324--333}.
\bibitem[{Humm et~al.(2003)Humm, Ballon, Hu, Ruan, Chui, Tulipano, Erdi,
  Koutcher, Zakian, Urano, Zanzonico, Mattis, Dyke, Chen, Harrington,
  O'Donoghue and Ling}]{humm2003stereotactic}
\bibinfo{author}{Humm, J.}, \bibinfo{author}{Ballon, D.}, \bibinfo{author}{Hu,
  Y.}, \bibinfo{author}{Ruan, S.}, \bibinfo{author}{Chui, C.},
  \bibinfo{author}{Tulipano, P.}, \bibinfo{author}{Erdi, A.},
  \bibinfo{author}{Koutcher, J.}, \bibinfo{author}{Zakian, K.},
  \bibinfo{author}{Urano, M.}, \bibinfo{author}{Zanzonico, P.},
  \bibinfo{author}{Mattis, C.}, \bibinfo{author}{Dyke, J.},
  \bibinfo{author}{Chen, Y.}, \bibinfo{author}{Harrington, P.},
  \bibinfo{author}{O'Donoghue, J.}, \bibinfo{author}{Ling, C.},
  \bibinfo{year}{2003}.
\newblock \bibinfo{title}{A stereotactic method for the three-dimensional
  registration of multi-modality biologic images in animals: {NMR}, {PET},
  histology, and autoradiography}.
\newblock \bibinfo{journal}{Medical physics} \bibinfo{volume}{30},
  \bibinfo{pages}{2303--2314}.
\bibitem[{Huynh et~al.(2016)Huynh, Gao, Kang, Wang, Zhang, Lian and
  Shen}]{huynh2016estimating}
\bibinfo{author}{Huynh, T.}, \bibinfo{author}{Gao, Y.}, \bibinfo{author}{Kang,
  J.}, \bibinfo{author}{Wang, L.}, \bibinfo{author}{Zhang, P.},
  \bibinfo{author}{Lian, J.}, \bibinfo{author}{Shen, D.}, \bibinfo{year}{2016}.
\newblock \bibinfo{title}{Estimating {CT} image from {MRI} data using
  structured random forest and auto-context model}.
\newblock \bibinfo{journal}{IEEE transactions on medical imaging}
  \bibinfo{volume}{35}, \bibinfo{pages}{174--183}.
\bibitem[{Iglesias et~al.(2015)Iglesias, Augustinack, Nguyen, Player, Player,
  Wright, Roy, Frosch, McKee, Wald, Fischl and
  Van~Leemput}]{iglesias2015computational}
\bibinfo{author}{Iglesias, J.E.}, \bibinfo{author}{Augustinack, J.C.},
  \bibinfo{author}{Nguyen, K.}, \bibinfo{author}{Player, C.M.},
  \bibinfo{author}{Player, A.}, \bibinfo{author}{Wright, M.},
  \bibinfo{author}{Roy, N.}, \bibinfo{author}{Frosch, M.P.},
  \bibinfo{author}{McKee, A.C.}, \bibinfo{author}{Wald, L.L.},
  \bibinfo{author}{Fischl, B.}, \bibinfo{author}{Van~Leemput, K.},
  \bibinfo{year}{2015}.
\newblock \bibinfo{title}{A computational atlas of the hippocampal formation
  using ex vivo, ultra-high resolution {MRI}: application to adaptive
  segmentation of in vivo {MRI}}.
\newblock \bibinfo{journal}{NeuroImage} \bibinfo{volume}{115},
  \bibinfo{pages}{117--137}.
\bibitem[{Iglesias et~al.(2017)Iglesias, Insausti, Lerma-Usabiaga, Van~Leemput,
  Ourselin, Fischl, Caballero-Gaudes and Paz-Alonso}]{iglesias2017thalamus}
\bibinfo{author}{Iglesias, J.E.}, \bibinfo{author}{Insausti, R.},
  \bibinfo{author}{Lerma-Usabiaga, G.}, \bibinfo{author}{Van~Leemput, K.},
  \bibinfo{author}{Ourselin, S.}, \bibinfo{author}{Fischl, B.},
  \bibinfo{author}{Caballero-Gaudes, C.}, \bibinfo{author}{Paz-Alonso, P.M.},
  \bibinfo{year}{2017}.
\newblock \bibinfo{title}{A probabilistic atlas of the thalamic nuclei
  combining ex vivo {MRI} and histology}.
\newblock \bibinfo{howpublished}{Organization for Human Brain Mapping (OHBM)}.
\bibitem[{Iglesias et~al.(2013)Iglesias, Konukoglu, Zikic, Glocker, Van~Leemput
  and Fischl}]{iglesias2013synthesizing}
\bibinfo{author}{Iglesias, J.E.}, \bibinfo{author}{Konukoglu, E.},
  \bibinfo{author}{Zikic, D.}, \bibinfo{author}{Glocker, B.},
  \bibinfo{author}{Van~Leemput, K.}, \bibinfo{author}{Fischl, B.},
  \bibinfo{year}{2013}.
\newblock \bibinfo{title}{Is synthesizing {MRI} contrast useful for
  inter-modality analysis?}, in: \bibinfo{booktitle}{International Conference
  on Medical Image Computing and Computer-Assisted Intervention},
  \bibinfo{organization}{Springer}. pp. \bibinfo{pages}{631--638}.
\bibitem[{Ju et~al.(2006)Ju, Warren, Carson, Bello, Kakadiaris, Chiu, Thaller
  and Eichele}]{ju20063d}
\bibinfo{author}{Ju, T.}, \bibinfo{author}{Warren, J.},
  \bibinfo{author}{Carson, J.}, \bibinfo{author}{Bello, M.},
  \bibinfo{author}{Kakadiaris, I.}, \bibinfo{author}{Chiu, W.},
  \bibinfo{author}{Thaller, C.}, \bibinfo{author}{Eichele, G.},
  \bibinfo{year}{2006}.
\newblock \bibinfo{title}{3d volume reconstruction of a mouse brain from
  histological sections using warp filtering}.
\newblock \bibinfo{journal}{Journal of Neuroscience Methods}
  \bibinfo{volume}{156}, \bibinfo{pages}{84--100}.
\bibitem[{Kim et~al.(1997)Kim, Boes, Frey and Meyer}]{kim1997mutual}
\bibinfo{author}{Kim, B.}, \bibinfo{author}{Boes, J.L.}, \bibinfo{author}{Frey,
  K.A.}, \bibinfo{author}{Meyer, C.R.}, \bibinfo{year}{1997}.
\newblock \bibinfo{title}{Mutual information for automated unwarping of rat
  brain autoradiographs}.
\newblock \bibinfo{journal}{Neuroimage} \bibinfo{volume}{5},
  \bibinfo{pages}{31--40}.
\bibitem[{Kim et~al.(2015)Kim, Glide-Hurst, Doemer, Wen, Movsas and
  Chetty}]{kim2015implementation}
\bibinfo{author}{Kim, J.}, \bibinfo{author}{Glide-Hurst, C.},
  \bibinfo{author}{Doemer, A.}, \bibinfo{author}{Wen, N.},
  \bibinfo{author}{Movsas, B.}, \bibinfo{author}{Chetty, I.J.},
  \bibinfo{year}{2015}.
\newblock \bibinfo{title}{Implementation of a novel algorithm for generating
  synthetic {CT} images from magnetic resonance imaging data sets for prostate
  cancer radiation therapy}.
\newblock \bibinfo{journal}{International Journal of Radiation Oncology*
  Biology* Physics} \bibinfo{volume}{91}, \bibinfo{pages}{39--47}.
\bibitem[{Krauth et~al.(2010)Krauth, Blanc, Poveda, Jeanmonod, Morel and
  Sz{\'e}kely}]{krauth2010mean}
\bibinfo{author}{Krauth, A.}, \bibinfo{author}{Blanc, R.},
  \bibinfo{author}{Poveda, A.}, \bibinfo{author}{Jeanmonod, D.},
  \bibinfo{author}{Morel, A.}, \bibinfo{author}{Sz{\'e}kely, G.},
  \bibinfo{year}{2010}.
\newblock \bibinfo{title}{A mean three-dimensional atlas of the human thalamus:
  generation from multiple histological data}.
\newblock \bibinfo{journal}{Neuroimage} \bibinfo{volume}{49},
  \bibinfo{pages}{2053--2062}.
\bibitem[{Maes et~al.(1997)Maes, Collignon, Vandermeulen, Marchal and
  Suetens}]{maes1997multimodality}
\bibinfo{author}{Maes, F.}, \bibinfo{author}{Collignon, A.},
  \bibinfo{author}{Vandermeulen, D.}, \bibinfo{author}{Marchal, G.},
  \bibinfo{author}{Suetens, P.}, \bibinfo{year}{1997}.
\newblock \bibinfo{title}{Multimodality image registration by maximization of
  mutual information}.
\newblock \bibinfo{journal}{IEEE transactions on medical imaging}
  \bibinfo{volume}{16}, \bibinfo{pages}{187--198}.
\bibitem[{Malandain et~al.(2004)Malandain, Bardinet, Nelissen and
  Vanduffel}]{malandain2004fusion}
\bibinfo{author}{Malandain, G.}, \bibinfo{author}{Bardinet, E.},
  \bibinfo{author}{Nelissen, K.}, \bibinfo{author}{Vanduffel, W.},
  \bibinfo{year}{2004}.
\newblock \bibinfo{title}{Fusion of autoradiographs with an mr volume using 2-d
  and 3-d linear transformations}.
\newblock \bibinfo{journal}{NeuroImage} \bibinfo{volume}{23},
  \bibinfo{pages}{111--127}.
\bibitem[{Mazziotta et~al.(2001)Mazziotta, Toga, Evans, Fox, Lancaster, Zilles,
  Woods, Paus, Simpson, Pike, Holmes, Collins, Thompson, MacDonald, Iacoboni,
  Schormann, Amunts, Palomero-Gallagher, Geyer, Parsons, Narr, Kabani,
  Le~Goualher, Boomsma, Cannon, Kawashima and
  Mazoyer}]{mazziotta2001probabilistic}
\bibinfo{author}{Mazziotta, J.}, \bibinfo{author}{Toga, A.},
  \bibinfo{author}{Evans, A.}, \bibinfo{author}{Fox, P.},
  \bibinfo{author}{Lancaster, J.}, \bibinfo{author}{Zilles, K.},
  \bibinfo{author}{Woods, R.}, \bibinfo{author}{Paus, T.},
  \bibinfo{author}{Simpson, G.}, \bibinfo{author}{Pike, B.},
  \bibinfo{author}{Holmes, C.}, \bibinfo{author}{Collins, L.},
  \bibinfo{author}{Thompson, P.}, \bibinfo{author}{MacDonald, D.},
  \bibinfo{author}{Iacoboni, M.}, \bibinfo{author}{Schormann, T.},
  \bibinfo{author}{Amunts, K.}, \bibinfo{author}{Palomero-Gallagher, N.},
  \bibinfo{author}{Geyer, S.}, \bibinfo{author}{Parsons, L.},
  \bibinfo{author}{Narr, K.}, \bibinfo{author}{Kabani, N.},
  \bibinfo{author}{Le~Goualher, G.}, \bibinfo{author}{Boomsma, D.},
  \bibinfo{author}{Cannon, T.}, \bibinfo{author}{Kawashima, R.},
  \bibinfo{author}{Mazoyer, B.}, \bibinfo{year}{2001}.
\newblock \bibinfo{title}{A probabilistic atlas and reference system for the
  human brain: International consortium for brain mapping ({ICBM})}.
\newblock \bibinfo{journal}{Philosophical Transactions of the Royal Society of
  London B: Biological Sciences} \bibinfo{volume}{356},
  \bibinfo{pages}{1293--1322}.
\bibitem[{Mazziotta et~al.(1995)Mazziotta, Toga, Evans, Fox and
  Lancaster}]{mazziotta1995probabilistic}
\bibinfo{author}{Mazziotta, J.C.}, \bibinfo{author}{Toga, A.W.},
  \bibinfo{author}{Evans, A.}, \bibinfo{author}{Fox, P.},
  \bibinfo{author}{Lancaster, J.}, \bibinfo{year}{1995}.
\newblock \bibinfo{title}{A probabilistic atlas of the human brain: Theory and
  rationale for its development: The international consortium for brain mapping
  ({ICBM})}.
\newblock \bibinfo{journal}{Neuroimage} \bibinfo{volume}{2},
  \bibinfo{pages}{89--101}.
\bibitem[{Modat et~al.(2014)Modat, Cash, Daga, Winston, Duncan and
  Ourselin}]{modat2014global}
\bibinfo{author}{Modat, M.}, \bibinfo{author}{Cash, D.M.},
  \bibinfo{author}{Daga, P.}, \bibinfo{author}{Winston, G.P.},
  \bibinfo{author}{Duncan, J.S.}, \bibinfo{author}{Ourselin, S.},
  \bibinfo{year}{2014}.
\newblock \bibinfo{title}{Global image registration using a symmetric
  block-matching approach}.
\newblock \bibinfo{journal}{Journal of Medical Imaging} \bibinfo{volume}{1},
  \bibinfo{pages}{024003--024003}.
\bibitem[{Modat et~al.(2010)Modat, Ridgway, Taylor, Lehmann, Barnes, Hawkes,
  Fox and Ourselin}]{modat2010fast}
\bibinfo{author}{Modat, M.}, \bibinfo{author}{Ridgway, G.R.},
  \bibinfo{author}{Taylor, Z.A.}, \bibinfo{author}{Lehmann, M.},
  \bibinfo{author}{Barnes, J.}, \bibinfo{author}{Hawkes, D.J.},
  \bibinfo{author}{Fox, N.C.}, \bibinfo{author}{Ourselin, S.},
  \bibinfo{year}{2010}.
\newblock \bibinfo{title}{Fast free-form deformation using graphics processing
  units}.
\newblock \bibinfo{journal}{Computer methods and programs in biomedicine}
  \bibinfo{volume}{98}, \bibinfo{pages}{278--284}.
\bibitem[{Moler and Van~Loan(2003)}]{moler2003nineteen}
\bibinfo{author}{Moler, C.}, \bibinfo{author}{Van~Loan, C.},
  \bibinfo{year}{2003}.
\newblock \bibinfo{title}{Nineteen dubious ways to compute the exponential of a
  matrix, twenty-five years later}.
\newblock \bibinfo{journal}{SIAM review} \bibinfo{volume}{45},
  \bibinfo{pages}{3--49}.
\bibitem[{Ourselin et~al.(2001)Ourselin, Roche, Subsol, Pennec and
  Ayache}]{ourselin2001reconstructing}
\bibinfo{author}{Ourselin, S.}, \bibinfo{author}{Roche, A.},
  \bibinfo{author}{Subsol, G.}, \bibinfo{author}{Pennec, X.},
  \bibinfo{author}{Ayache, N.}, \bibinfo{year}{2001}.
\newblock \bibinfo{title}{Reconstructing a 3d structure from serial
  histological sections}.
\newblock \bibinfo{journal}{Image and vision computing} \bibinfo{volume}{19},
  \bibinfo{pages}{25--31}.
\bibitem[{Parisi(1988)}]{parisi1988statistical}
\bibinfo{author}{Parisi, G.}, \bibinfo{year}{1988}.
\newblock \bibinfo{title}{Statistical field theory}.
\newblock \bibinfo{publisher}{Addison-Wesley}.
\bibitem[{Pitiot et~al.(2006)Pitiot, Bardinet, Thompson and
  Malandain}]{pitiot2006piecewise}
\bibinfo{author}{Pitiot, A.}, \bibinfo{author}{Bardinet, E.},
  \bibinfo{author}{Thompson, P.M.}, \bibinfo{author}{Malandain, G.},
  \bibinfo{year}{2006}.
\newblock \bibinfo{title}{Piecewise affine registration of biological images
  for volume reconstruction}.
\newblock \bibinfo{journal}{Medical image analysis} \bibinfo{volume}{10},
  \bibinfo{pages}{465--483}.
\bibitem[{Pluim et~al.(2003)Pluim, Maintz and Viergever}]{pluim2003mutual}
\bibinfo{author}{Pluim, J.P.}, \bibinfo{author}{Maintz, J.A.},
  \bibinfo{author}{Viergever, M.A.}, \bibinfo{year}{2003}.
\newblock \bibinfo{title}{Mutual-information-based registration of medical
  images: a survey}.
\newblock \bibinfo{journal}{IEEE transactions on medical imaging}
  \bibinfo{volume}{22}, \bibinfo{pages}{986--1004}.
\bibitem[{Rangarajan et~al.(1997)Rangarajan, Chui, Mjolsness, Pappu, Davachi,
  Goldman-Rakic and Duncan}]{rangarajan1997robust}
\bibinfo{author}{Rangarajan, A.}, \bibinfo{author}{Chui, H.},
  \bibinfo{author}{Mjolsness, E.}, \bibinfo{author}{Pappu, S.},
  \bibinfo{author}{Davachi, L.}, \bibinfo{author}{Goldman-Rakic, P.},
  \bibinfo{author}{Duncan, J.}, \bibinfo{year}{1997}.
\newblock \bibinfo{title}{A robust point-matching algorithm for autoradiograph
  alignment}.
\newblock \bibinfo{journal}{Medical image analysis} \bibinfo{volume}{1},
  \bibinfo{pages}{379--398}.
\bibitem[{Roy et~al.(2014)Roy, Carass, Jog, Prince and Lee}]{roy2014mr}
\bibinfo{author}{Roy, S.}, \bibinfo{author}{Carass, A.}, \bibinfo{author}{Jog,
  A.}, \bibinfo{author}{Prince, J.L.}, \bibinfo{author}{Lee, J.},
  \bibinfo{year}{2014}.
\newblock \bibinfo{title}{{MR} to {CT} registration of brains using image
  synthesis}, in: \bibinfo{booktitle}{Proceedings of SPIE},
  \bibinfo{organization}{NIH Public Access}.
\bibitem[{Roy et~al.(2013)Roy, Carass and Prince}]{roy2013magnetic}
\bibinfo{author}{Roy, S.}, \bibinfo{author}{Carass, A.},
  \bibinfo{author}{Prince, J.L.}, \bibinfo{year}{2013}.
\newblock \bibinfo{title}{Magnetic resonance image example-based contrast
  synthesis}.
\newblock \bibinfo{journal}{IEEE transactions on medical imaging}
  \bibinfo{volume}{32}, \bibinfo{pages}{2348--2363}.
\bibitem[{Rueckert et~al.(1999)Rueckert, Sonoda, Hayes, Hill, Leach and
  Hawkes}]{rueckert1999nonrigid}
\bibinfo{author}{Rueckert, D.}, \bibinfo{author}{Sonoda, L.I.},
  \bibinfo{author}{Hayes, C.}, \bibinfo{author}{Hill, D.L.},
  \bibinfo{author}{Leach, M.O.}, \bibinfo{author}{Hawkes, D.J.},
  \bibinfo{year}{1999}.
\newblock \bibinfo{title}{Nonrigid registration using free-form deformations:
  application to breast {MR} images}.
\newblock \bibinfo{journal}{IEEE transactions on medical imaging}
  \bibinfo{volume}{18}, \bibinfo{pages}{712--721}.
\bibitem[{Saygin et~al.(2017)Saygin, Kliemann, Iglesias, van~der Kouwe, Boyd,
  Reuter, Stevens, Van~Leemput, McKee, Frosch, Fischl and
  Augustinack}]{saygin2017high}
\bibinfo{author}{Saygin, Z.}, \bibinfo{author}{Kliemann, D.},
  \bibinfo{author}{Iglesias, J.}, \bibinfo{author}{van~der Kouwe, A.},
  \bibinfo{author}{Boyd, E.}, \bibinfo{author}{Reuter, M.},
  \bibinfo{author}{Stevens, A.}, \bibinfo{author}{Van~Leemput, K.},
  \bibinfo{author}{McKee, A.}, \bibinfo{author}{Frosch, M.},
  \bibinfo{author}{Fischl, B.}, \bibinfo{author}{Augustinack, J.C.},
  \bibinfo{year}{2017}.
\newblock \bibinfo{title}{High-resolution magnetic resonance imaging reveals
  nuclei of the human amygdala: manual segmentation to automatic atlas}.
\newblock \bibinfo{journal}{NeuroImage} \bibinfo{volume}{155},
  \bibinfo{pages}{370--382}.
\bibitem[{Schmitt et~al.(2007)Schmitt, Modersitzki, Heldmann, Wirtz and
  Fischer}]{schmitt2007image}
\bibinfo{author}{Schmitt, O.}, \bibinfo{author}{Modersitzki, J.},
  \bibinfo{author}{Heldmann, S.}, \bibinfo{author}{Wirtz, S.},
  \bibinfo{author}{Fischer, B.}, \bibinfo{year}{2007}.
\newblock \bibinfo{title}{Image registration of sectioned brains}.
\newblock \bibinfo{journal}{International Journal of Computer Vision}
  \bibinfo{volume}{73}, \bibinfo{pages}{5--39}.
\bibitem[{Shattuck et~al.(2008)Shattuck, Mirza, Adisetiyo, Hojatkashani,
  Salamon, Narr, Poldrack, Bilder and Toga}]{shattuck2008construction}
\bibinfo{author}{Shattuck, D.W.}, \bibinfo{author}{Mirza, M.},
  \bibinfo{author}{Adisetiyo, V.}, \bibinfo{author}{Hojatkashani, C.},
  \bibinfo{author}{Salamon, G.}, \bibinfo{author}{Narr, K.L.},
  \bibinfo{author}{Poldrack, R.A.}, \bibinfo{author}{Bilder, R.M.},
  \bibinfo{author}{Toga, A.W.}, \bibinfo{year}{2008}.
\newblock \bibinfo{title}{Construction of a {3D} probabilistic atlas of human
  cortical structures}.
\newblock \bibinfo{journal}{Neuroimage} \bibinfo{volume}{39},
  \bibinfo{pages}{1064--1080}.
\bibitem[{Siversson et~al.(2015)Siversson, Nordstr{\"o}m, Nilsson, Nyholm,
  Jonsson, Gunnlaugsson and Olsson}]{siversson2015mri}
\bibinfo{author}{Siversson, C.}, \bibinfo{author}{Nordstr{\"o}m, F.},
  \bibinfo{author}{Nilsson, T.}, \bibinfo{author}{Nyholm, T.},
  \bibinfo{author}{Jonsson, J.}, \bibinfo{author}{Gunnlaugsson, A.},
  \bibinfo{author}{Olsson, L.E.}, \bibinfo{year}{2015}.
\newblock \bibinfo{title}{{MRI} only prostate radiotherapy planning using the
  statistical decomposition algorithm}.
\newblock \bibinfo{journal}{Medical physics} \bibinfo{volume}{42},
  \bibinfo{pages}{6090--6097}.
\bibitem[{Tu et~al.(2008)Tu, Narr, Doll{\'a}r, Dinov, Thompson and
  Toga}]{tu2008brain}
\bibinfo{author}{Tu, Z.}, \bibinfo{author}{Narr, K.L.},
  \bibinfo{author}{Doll{\'a}r, P.}, \bibinfo{author}{Dinov, I.},
  \bibinfo{author}{Thompson, P.M.}, \bibinfo{author}{Toga, A.W.},
  \bibinfo{year}{2008}.
\newblock \bibinfo{title}{Brain anatomical structure segmentation by hybrid
  discriminative/generative models}.
\newblock \bibinfo{journal}{IEEE transactions on medical imaging}
  \bibinfo{volume}{27}, \bibinfo{pages}{495--508}.
\bibitem[{van Tulder and de~Bruijne(2015)}]{van2015does}
\bibinfo{author}{van Tulder, G.}, \bibinfo{author}{de~Bruijne, M.},
  \bibinfo{year}{2015}.
\newblock \bibinfo{title}{Why does synthesized data improve multi-sequence
  classification?}, in: \bibinfo{booktitle}{International Conference on Medical
  Image Computing and Computer-Assisted Intervention},
  \bibinfo{organization}{Springer}. pp. \bibinfo{pages}{531--538}.
\bibitem[{Van~Leemput et~al.(1999)Van~Leemput, Maes, Vandermeulen and
  Suetens}]{van1999automated}
\bibinfo{author}{Van~Leemput, K.}, \bibinfo{author}{Maes, F.},
  \bibinfo{author}{Vandermeulen, D.}, \bibinfo{author}{Suetens, P.},
  \bibinfo{year}{1999}.
\newblock \bibinfo{title}{Automated model-based bias field correction of mr
  images of the brain}.
\newblock \bibinfo{journal}{IEEE transactions on medical imaging}
  \bibinfo{volume}{18}, \bibinfo{pages}{885--896}.
\bibitem[{Vercauteren et~al.(2007)Vercauteren, Pennec, Perchant and
  Ayache}]{vercauteren2007non}
\bibinfo{author}{Vercauteren, T.}, \bibinfo{author}{Pennec, X.},
  \bibinfo{author}{Perchant, A.}, \bibinfo{author}{Ayache, N.},
  \bibinfo{year}{2007}.
\newblock \bibinfo{title}{Non-parametric diffeomorphic image registration with
  the demons algorithm}.
\newblock \bibinfo{journal}{Medical Image Computing and Computer-Assisted
  Intervention--MICCAI 2007} , \bibinfo{pages}{319--326}.
\bibitem[{Wang et~al.(2017)Wang, Li, Zuluaga, Pratt, Patel, Aertsen, Doel,
  David, Deprest, Ourselin and Vercauteren}]{wang2017interactive}
\bibinfo{author}{Wang, G.}, \bibinfo{author}{Li, W.}, \bibinfo{author}{Zuluaga,
  M.A.}, \bibinfo{author}{Pratt, R.}, \bibinfo{author}{Patel, P.A.},
  \bibinfo{author}{Aertsen, M.}, \bibinfo{author}{Doel, T.},
  \bibinfo{author}{David, A.L.}, \bibinfo{author}{Deprest, J.},
  \bibinfo{author}{Ourselin, S.}, \bibinfo{author}{Vercauteren, T.},
  \bibinfo{year}{2017}.
\newblock \bibinfo{title}{Interactive medical image segmentation using deep
  learning with image-specific fine-tuning}.
\newblock \bibinfo{journal}{arXiv preprint arXiv:1710.04043} .
\bibitem[{Wells et~al.(1996)Wells, Viola, Atsumi, Nakajima and
  Kikinis}]{wells1996multi}
\bibinfo{author}{Wells, W.M.}, \bibinfo{author}{Viola, P.},
  \bibinfo{author}{Atsumi, H.}, \bibinfo{author}{Nakajima, S.},
  \bibinfo{author}{Kikinis, R.}, \bibinfo{year}{1996}.
\newblock \bibinfo{title}{Multi-modal volume registration by maximization of
  mutual information}.
\newblock \bibinfo{journal}{Medical image analysis} \bibinfo{volume}{1},
  \bibinfo{pages}{35--51}.
\bibitem[{Wolterink et~al.(2017)Wolterink, Dinkla, Savenije, Seevinck, van~den
  Berg and I{\v{s}}gum}]{wolterink2017deep}
\bibinfo{author}{Wolterink, J.M.}, \bibinfo{author}{Dinkla, A.M.},
  \bibinfo{author}{Savenije, M.H.}, \bibinfo{author}{Seevinck, P.R.},
  \bibinfo{author}{van~den Berg, C.A.}, \bibinfo{author}{I{\v{s}}gum, I.},
  \bibinfo{year}{2017}.
\newblock \bibinfo{title}{Deep {MR} to {CT} synthesis using unpaired data}, in:
  \bibinfo{booktitle}{International Workshop on Simulation and Synthesis in
  Medical Imaging}, \bibinfo{organization}{Springer}. pp.
  \bibinfo{pages}{14--23}.
\bibitem[{Yang et~al.(2012)Yang, Richards, Kurniawan, Petrou and
  Reutens}]{yang2012mri}
\bibinfo{author}{Yang, Z.}, \bibinfo{author}{Richards, K.},
  \bibinfo{author}{Kurniawan, N.D.}, \bibinfo{author}{Petrou, S.},
  \bibinfo{author}{Reutens, D.C.}, \bibinfo{year}{2012}.
\newblock \bibinfo{title}{Mri-guided volume reconstruction of mouse brain from
  histological sections}.
\newblock \bibinfo{journal}{Journal of neuroscience methods}
  \bibinfo{volume}{211}, \bibinfo{pages}{210--217}.
\bibitem[{Yushkevich et~al.(2006)Yushkevich, Avants, Ng, Hawrylycz, Burstein,
  Zhang and Gee}]{yushkevich20063d}
\bibinfo{author}{Yushkevich, P.A.}, \bibinfo{author}{Avants, B.B.},
  \bibinfo{author}{Ng, L.}, \bibinfo{author}{Hawrylycz, M.},
  \bibinfo{author}{Burstein, P.D.}, \bibinfo{author}{Zhang, H.},
  \bibinfo{author}{Gee, J.C.}, \bibinfo{year}{2006}.
\newblock \bibinfo{title}{3d mouse brain reconstruction from histology using a
  coarse-to-fine approach}.
\newblock \bibinfo{journal}{Lecture Notes in Computer Science}
  \bibinfo{volume}{4057}, \bibinfo{pages}{230--237}.
\bibitem[{Yushkevich et~al.(2009)Yushkevich, Avants, Pluta, Das, Minkoff,
  Mechanic-Hamilton, Glynn, Pickup, Liu, Gee, Grossman and
  Detre}]{yushkevich2009high}
\bibinfo{author}{Yushkevich, P.A.}, \bibinfo{author}{Avants, B.B.},
  \bibinfo{author}{Pluta, J.}, \bibinfo{author}{Das, S.},
  \bibinfo{author}{Minkoff, D.}, \bibinfo{author}{Mechanic-Hamilton, D.},
  \bibinfo{author}{Glynn, S.}, \bibinfo{author}{Pickup, S.},
  \bibinfo{author}{Liu, W.}, \bibinfo{author}{Gee, J.C.},
  \bibinfo{author}{Grossman, M.}, \bibinfo{author}{Detre, J.A.},
  \bibinfo{year}{2009}.
\newblock \bibinfo{title}{A high-resolution computational atlas of the human
  hippocampus from postmortem magnetic resonance imaging at {9.4T }}.
\newblock \bibinfo{journal}{Neuroimage} \bibinfo{volume}{44},
  \bibinfo{pages}{385--398}.
\bibitem[{Zhao et~al.(1993)Zhao, Young and Ginsberg}]{zhao1993registration}
\bibinfo{author}{Zhao, W.}, \bibinfo{author}{Young, T.Y.},
  \bibinfo{author}{Ginsberg, M.D.}, \bibinfo{year}{1993}.
\newblock \bibinfo{title}{Registration and three-dimensional reconstruction of
  autoradiographic images by the disparity analysis method}.
\newblock \bibinfo{journal}{IEEE Transactions on medical imaging}
  \bibinfo{volume}{12}, \bibinfo{pages}{782--791}.
\bibitem[{Zhu et~al.(2017)Zhu, Park, Isola and Efros}]{CycleGAN2017}
\bibinfo{author}{Zhu, J.Y.}, \bibinfo{author}{Park, T.},
  \bibinfo{author}{Isola, P.}, \bibinfo{author}{Efros, A.A.},
  \bibinfo{year}{2017}.
\newblock \bibinfo{title}{Unpaired image-to-image translation using
  cycle-consistent adversarial networks}.
\newblock \bibinfo{journal}{arXiv preprint arXiv:1703.10593} .

\end{thebibliography}

\end{document}